# High Dimensional Semiparametric Latent Graphical Model for Mixed Data

Jianqing Fan* Han Liu† Yang Ning‡ Hui Zou§

**Abstract**

Graphical models are commonly used tools for modeling multivariate random variables. While there exist many convenient multivariate distributions such as Gaussian distribution for continuous data, mixed data with the presence of discrete variables or a combination of both continuous and discrete variables poses new challenges in statistical modeling. In this paper, we propose a semiparametric model named latent Gaussian copula model for binary and mixed data. The observed binary data are assumed to be obtained by dichotomizing a latent variable satisfying the Gaussian copula distribution or the nonparanormal distribution. The latent Gaussian model with the assumption that the latent variables are multivariate Gaussian is a special case of the proposed model. A novel rank-based approach is proposed for both latent graph estimation and latent principal component analysis. Theoretically, the proposed methods achieve the same rates of convergence for both precision matrix estimation and eigenvector estimation, as if the latent variables were observed. Under similar conditions, the consistency of graph structure recovery and feature selection for leading eigenvectors is established. The performance of the proposed methods is numerically assessed through simulation studies, and the usage of our methods is illustrated by a genetic dataset.

**Keyword:** Discrete data, Gaussian copula, Latent variable, Mixed data, Nonparanormal, Principal component analysis, Rank-based statistic

## 1 Introduction

Graphical models (Lauritzen, 1996) have been widely used to explore the dependence structure of multivariate distributions, arising in many research areas including machine learning, image analysis, statistical physics and epidemiology. In these applications, the data collected often have high dimensionality and relatively low sample size. Under this high dimensional setting, parameter estimation and edge structure learning in the graphical model attracts increasing attention in statistics.

*Department of Operations Research Research and Financial Engineering, Princeton University, Princeton, NJ 08544, USA; e-mail: jqfan@princeton.edu.

†Department of Operations Research Research and Financial Engineering, Princeton University, Princeton, NJ 08544, USA; e-mail: hanliu@princeton.edu.

‡Department of Statistics and Actuarial Science, University of Waterloo, Waterloo, Canada; e-mail: y4ning@uwaterloo.ca.

§School of Statistics, University of Minnesota, Minneapolis, MN 55455, USA; e-mail: zouxx019@umn.edu.

Due to the mathematical simplicity and wide applicability, Gaussian graphical models have been extensively studied by Meinshausen and Bühlmann (2006); Yuan and Lin (2007); Rothman et al. (2008); Friedman et al. (2008); d'Aspremont et al. (2008); Fan et al. (2009); Lam and Fan (2009); Yuan (2010); Cai et al. (2011), among others. To relax the Gaussian distribution assumption, Xue and Zou (2012); Liu et al. (2009, 2012) proposed a semiparametric Gaussian copula model for modeling continuous data by allowing for monotonic univariate transformations.

Both the Gaussian and Gaussian copula models are only tailored for modeling continuous data. However, many multivariate problems may contain discrete data or data of hybrid types with both discrete and continuous variables. For instance, the genomic data such as the RNA-seq data produced by modern high-throughput sequencing technologies are usually count data. In social science, the covariate information collected by sample survey often contains both continuous and discrete variables. For binary data, Xue et al. (2012); Höfling and Tibshirani (2009); Ravikumar et al. (2010) proposed a penalized pseudo-likelihood approach under the Ising model. To handle more general non-Gaussian data such as the count data, Yang et al. (2013) proposed the class of exponential family graphical models, which can be viewed as a multivariate extension of the exponential family. Loh et al. (2013) studied the correspondence between conditional independence and the structure of the inverse covariance matrix in discrete graphs. Recently, the conditional Gaussian distribution is used by Lee and Hastie (2012); Cheng et al. (2013); Fellinghauer et al. (2013) to model the mixed data.

While the graphical model can be used to explore the dependence structure among multivariate random variables, it requires other techniques for finding patterns in high dimensional data. One such procedure is the principal component analysis (PCA), which has been used extensively in applications such as face recognition and image compression (Jolliffe, 2005). In the high dimensional setting, it is a common assumption that the leading eigenvectors are sparse. To utilize the sparsity assumption, different types of sparse PCA have been developed; see d'Aspremont et al. (2004); Zou et al. (2006); Shen and Huang (2008); Witten et al. (2009); Journée et al. (2010); Zhang and El Ghaoui (2011); Fan et al. (2013). The theoretical properties of sparse PCA for parameter estimation and feature selection have been studied by Amini and Wainwright (2009), Johnstone and Lu (2009), Ma (2013), Paul and Johnstone (2012), Vu and Lei (2012), and Berthet and Rigollet (2012), among others. While the sparse PCA methods are developed extensively for Gaussian data, little has been done for high dimensional discrete data.

In many applications, it is often reasonable to assume that the discrete variable is obtained by discretizing a latent (unobserved) variable (Skrondal and Rabe-Hesketh, 2007). For instance, in psychology, the latent variables can represent abstract concepts such as human feeling or recognition that exist in hidden form but are not directly measurable, and instead, they can be measured indirectly by some surrogate variables. Motivated by the latent variable formulation, we propose a generative model named the latent Gaussian copula model for binary and mixed data. We assume that the observed discrete data are generated by discretizing a latent continuous variable at some unknown cutoff. Our model is semiparametric in the sense that the latent variables satisfy the Gaussian copula distribution (Xue and Zou, 2012) or the nonparanormal distribution (Liu et al., 2009, 2012). The proposed model reduces to the latent Gaussian model (Han and Pan, 2012) when the latent variables are multivariate Gaussian. We find that the family of latent Gaussian copula model is equivalent to the latent Gaussian model for binary data, and is strictly larger than the latent Gaussian model for mixed data. In the latent model framework, the latent variables usually represent unobserved quantifies of interest, such as human feeling or recognition in psychology. Hence, our goal is to infer the conditional independence structure among latent variables, which provides deeper understanding of the unknown mechanism than that among the observed variables. Thus, our model complements the existing work on high dimensional discrete or mixed graphical models, which mostly focused on learning conditional independence among observed variables.

To make our investigation more comprehensive, we consider both latent graph estimation and latent principal component analysis. Under the latent Gaussian copula model, the graph structure is characterized by the sparsity pattern of the precision matrix. However, due to high computational cost, the penalized maximum likelihood estimation cannot be used to estimate the precision matrix. Instead, we propose a novel rank-based approach for the observed binary or mixed data, which can be viewed as a nontrivial extension of the method developed by Xue and Zou (2012); Liu et al. (2012) for the nonparanormal model. Similarly, the rank-based approach can be applied to conduct sparse PCA for latent variables. A two-stage method is used to compute the scale-invariant sparse eigenvectors. Theoretically, we show that the proposed methods achieve the same rates of convergence for both graph estimation and eigenvector estimation, as if the latent variables were observed. Under similar conditions, the consistency of graph structure recovery and feature selection for leading eigenvectors is established. To the best of our knowledge, our work provides the first method for sparse PCA for binary or mixed data with theoretical guarantees under the high dimensional scaling.

The rest of the article is organized as follows. In Section 2, we review the Gaussian copula model. In Section 3, we introduce the latent Gaussian copula model for binary data and a rank-based estimator of latent correlation matrix. We consider latent graph estimation and latent scale-invariant PCA in Sections 4 and 5, respectively. In Section 6, we extend our procedures to the mixed data. We conduct extensive simulation studies and apply our methods to a real data example in Section 7. Discussion and concluding remarks are presented in the last section. Technical derivations and proofs are deferred to the appendix.

# 2 Background

## 2.1 Notation

For the following development, we introduce some notation. Let $\mathbf{M} = [\mathbf{M}_{jk}] \in \mathbb{R}^{d\times d}$ and $\boldsymbol{v} = (v_1, ..., v_d)^T \in \mathbb{R}^d$ be a $d$-dimensional matrix and a $d$-dimensional vector. We denote $\boldsymbol{v}_I$ to be the subvector of $\boldsymbol{v}$ whose entries are indexed by a set $I$. We also denote $\mathbf{M}_{I,J}$ to be the submatrix of $\mathbf{M}$ whose rows are indexed by $I$ and columns are indexed by $J$. Let $\mathbf{M}_{I*}$ and $\mathbf{M}_{*J}$ be the submatrix of $\mathbf{M}$ with rows in $I$, and the submatrix of $\mathbf{M}$ with columns in $J$. Let $\text{supp}(\boldsymbol{v}) := \{j : v_j \neq 0\}$. For $0 < q < \infty$, we define the $\ell_0$, $\ell_q$ and $\ell_\infty$ vector norms as $||\boldsymbol{v}||_0 := \text{card}(\text{supp}(\boldsymbol{v}))$, $||\boldsymbol{v}||_q := (\sum_{i=1}^d |v_i|^q)^{1/q}$ and $||\boldsymbol{v}||_\infty := \max_{1\le i\le d} |v_i|$. We define $||\mathbf{M}||_{\max} := \max\{|\mathbf{M}_{ij}|\}$ as the matrix $\ell_{\max}$ norm, $||\mathbf{M}||_2$ as the spectral norm and $||\mathbf{M}||_F$ as the Frobenius norm. Let $||\mathbf{M}||_{a,b}$ denote the $\ell_b$ norm of the vector of rowwise $\ell_a$ norms of $\mathbf{M}$. Let $\Lambda_j(\mathbf{M})$ be the $j$-th largest eigenvalue of $\mathbf{M}$. In particular, we denote $\Lambda_{\min}(\mathbf{M}) := \Lambda_d(\mathbf{M})$ and $\Lambda_{\max}(\mathbf{M}) := \Lambda_1(\mathbf{M})$ to be the smallest and largest eigenvalues of $\mathbf{M}$. We define $\text{vec}(\mathbf{M}) := (\mathbf{M}_{*1}^T, \ldots, \mathbf{M}_{*d}^T)^T$ and $\mathbb{S}^{d-1} := \{\boldsymbol{v} \in \mathbb{R}^d : ||\boldsymbol{v}||_2 = 1\}$ be the $d$-dimensional unit sphere. For any two vectors $\boldsymbol{a}, \boldsymbol{b} \in \mathbb{R}^d$ and any two squared matrices $\mathbf{A}, \mathbf{B} \in \mathbb{R}^{d\times d}$, we denote the inner product of $\boldsymbol{a}$ and $\boldsymbol{b}$, $\mathbf{A}$ and $\mathbf{B}$ by $\langle \boldsymbol{a}, \boldsymbol{b}\rangle := \boldsymbol{a}^T\boldsymbol{b}$ and $\langle \mathbf{A}, \mathbf{B}\rangle := \mathrm{Tr}(\mathbf{A}^T\mathbf{B})$. For any matrix $\mathbf{M} \in \mathbb{R}^{d\times d}$, we denote $\mathrm{diag}(\mathbf{M})$ to be the diagonal matrix with the same diagonal entries as $\mathbf{M}$. Let $\mathbf{I}_d$ be the $d$ by $d$ identity matrix.

### 2.2 Gaussian Copula Model

In multivariate analysis, the Gaussian model is commonly used due to its mathematical simplicity (Lauritzen, 1996). Although the Gaussian model has been widely applied, the normality assumption is rather restrictive. To relax this assumption, Xue and Zou (2012); Liu et al. (2009, 2012) proposed a semiparametric Gaussian copula model, also called nonparanormal model.

**Definition 2.1 (Gaussian copula model or nonparanormal model).** A random vector $\mathbf{X} = (X_1, ..., X_d)^T$ is sampled from the Gaussian copula model or nonparanormal model, if and only if there exists a set of monotonically increasing transformations $f = (f_j)_{j=1}^d$, satisfying $f(\mathbf{X}) = \{f_1(X_1), ..., f_d(X_d)\}^T \sim \mathrm{N}_d(0, \mathbf{\Sigma})$ with $\mathrm{diag}(\mathbf{\Sigma}) = \mathbf{I}_d$. Then we denote $\mathbf{X} \sim \mathrm{NPN}(0, \mathbf{\Sigma}, f)$.

Under the Gaussian copula model, the sparsity pattern of $\mathbf{\Omega} = \mathbf{\Sigma}^{-1}$ encodes the conditional independence among $\mathbf{X}$. Specifically, $X_i$ and $X_j$ are independent given the remaining variables $\mathbf{X}_{-(i,j)}$ if and only if $\Omega_{ij} = 0$. Hence, inferring the graph structure under the Gaussian copula model can be accomplished by estimating $\mathbf{\Omega}$.

Given $n$ independent copies of random vector $\mathbf{X}$, an estimation procedure based on the rank-based correlations such as Spearman's rho and Kendall's tau can be used to estimate $\mathbf{\Sigma}$ (Xue and Zou, 2012; Liu et al., 2012). Such an approach has the advantage of avoiding the estimation of the marginal transformations $(f_j)_{j=1}^d$. Once an estimator of $\mathbf{\Sigma}$ is derived, it can be plugged into any precision matrix estimation procedure developed for the Gaussian graphical model (Yuan, 2010; Cai et al., 2011; Friedman et al., 2008). Liu et al. (2012) and Xue and Zou (2012) showed that such a procedure achieves the optimal parametric rates for parameter estimation and graph recovery.

## 3 Latent Gaussian Copula Model for Binary Data

### 3.1 The model

Let $\mathbf{X} = (X_1, \ldots, X_d)^T \in \{0, 1\}^d$ be a $d$-dimensional 0/1-random vector. The latent Gaussian copula model for binary data is defined as follows.

**Definition 3.1 (Latent Gaussian copula model for binary data).** The 0/1-random vector $\mathbf{X}$ satisfies the latent Gaussian copula model, if there exists a $d$-dimensional random vector $\boldsymbol{Z} = (Z_1, ..., Z_d)^T \sim \mathrm{NPN}(0, \mathbf{\Sigma}, f)$ such that

$$X_j = I(Z_j > C_j) \quad \text{for all } j = 1, \ldots, d,$$

where $I(\cdot)$ is the indicator function and $\boldsymbol{C} = (C_1, ..., C_d)$ is a vector of constants. Then we denote $\mathbf{X} \sim \mathrm{LNPN}(0, \mathbf{\Sigma}, f, \boldsymbol{C})$. We call $\mathbf{\Sigma}$ the latent correlation matrix. When $\boldsymbol{Z} \sim \mathrm{N}(0, \mathbf{\Sigma})$, we say $\mathbf{X}$ satisfies the latent Gaussian model $\mathrm{LN}(0, \mathbf{\Sigma}, \boldsymbol{C})$.

In the latent Gaussian copula model, the 0/1-random vector $\mathbf{X}$ is generated by a latent continuous random vector $\boldsymbol{Z}$ truncated at unknown constants $\boldsymbol{C}$. Due to the flexibility of the Gaussian copula model, we allow the distribution of the latent variable $\boldsymbol{Z}$ to be skewed or multimodal. Let $\boldsymbol{\Omega} = \boldsymbol{\Sigma}^{-1}$ denote the latent precision matrix. By Liu et al. (2009), we know that the zero pattern of $\boldsymbol{\Omega}$ characterizes the conditional independence among the latent variables $\boldsymbol{Z}$.

Let $f(z_1, ..., z_d)$ denote the probability density function of $\boldsymbol{Z}$. It is easily seen that the joint probability mass function of $\mathbf{X}$ at point $\mathbf{x} \in \{0,1\}^d$ is given by

$$\mathbb{P}(\mathbf{X} = \mathbf{x}; \boldsymbol{C}, \boldsymbol{\Sigma}, f) = \frac{1}{(2\pi)^{d/2}|\boldsymbol{\Sigma}|^{1/2}} \int_{R(\boldsymbol{u})} \exp\Big\{-\frac{1}{2}\boldsymbol{u}^T\boldsymbol{\Sigma}^{-1}\boldsymbol{u}\Big\} du_1...du_d, \quad (1)$$

where $\boldsymbol{u} = (u_1, ..., u_d)$ and the integration region is $R(\boldsymbol{u}) = R_1 \times ... \times R_d$ with $R_j = [f_j(C_j), \infty]$ if $x_j = 1$ and $R_{2j} = [-\infty, f_j(C_j)]$ otherwise for $j = 1, ...d$,

Note that the latent Gaussian copula model in Definition 3.1 involves parameters $(\boldsymbol{\Sigma}, f, \boldsymbol{C})$. Merely based on the binary random vector $\mathbf{X}$, certain parameters in the model are not identifiable. By (1), we find that only $f_j(C_j)$, $j = 1, ..., d$, are identifiable. Denote $\boldsymbol{\Delta} = (\Delta_1, ..., \Delta_d)$, where $\Delta_j = f_j(C_j)$. For notational simplicity, we write $\mathrm{LNPN}(\boldsymbol{\Sigma}, \boldsymbol{\Delta})$ for $\mathrm{LNPN}(0, \boldsymbol{\Sigma}, f, \boldsymbol{C})$.

Another consequence of the identifiability constraint is that, the proposed latent Gaussian copula model is equivalent to the latent Gaussian model for binary outcomes. This phenomenon is expected, because the binary outcomes contain little information to identify the marginal transformations, whose effect can be offset by properly shifting the cutoff constants in the latent Gaussian model for binary outcomes. However, in Section 6, it is seen that the family of latent Gaussian copula model is strictly larger than the latent Gaussian model, when the observed variable $\mathbf{X}$ has both continuous and discrete components.

### 3.2 Rank Based Latent Correlation Matrix Estimator

Assume that we observe $n$ independent binary vector-valued data $\mathbf{X}_1, \ldots, \mathbf{X}_n \sim \mathrm{LNPN}(\boldsymbol{\Sigma}, \boldsymbol{\Delta})$. In this section, we propose a convenient rank based estimator of $\boldsymbol{\Sigma}$. Due to the fact that the latent variable $\boldsymbol{Z}$ is not observed, one cannot directly use the method in Xue and Zou (2012); Liu et al. (2012) to estimate $\boldsymbol{\Sigma}$. Instead, our idea is based on the following Kendall's tau calculated from the observed binary data $(X_{1j}, X_{1k}), ..., (X_{nj}, X_{nk})$,

$$\begin{aligned}\widehat{\tau}_{jk} &= \frac{2}{n(n-1)} \sum_{1 \le i < i' \le n} \left(X_{ij} - X_{i'j}\right)\left(X_{ik} - X_{i'k}\right) \\ &= \frac{2(n_a n_d - n_b n_c)}{n(n-1)},\end{aligned} \quad (2)$$

where we utilize the fact that $\mathrm{sign}(X_{ij} - X_{i'j}) = X_{ij} - X_{i'j}$ for binary variables, and $n_a, n_b, n_c$ and $n_d$ are given by the $2 \times 2$ contingency table with

$$n_a = \sum_{i=1}^n I(X_{ij} = 1 \text{ and } X_{ik} = 1), \quad n_b = \sum_{i=1}^n I(X_{ij} = 1 \text{ and } X_{ik} = 0),$$
$$n_c = \sum_{i=1}^n I(X_{ij} = 0 \text{ and } X_{ik} = 1), \quad n_d = \sum_{i=1}^n I(X_{ij} = 0 \text{ and } X_{ik} = 0),$$

with the dependence on $j$ and $k$ suppressed. Let $\Phi(\cdot)$ be the cumulative distribution function of the standard normal distribution and

$$\Phi_2(u, v, t) = \int_{x_1 < u} \int_{x_2 < v} \phi_2(x_1, x_2; t) dx_1 dx_2$$

be the cumulative distribution function of the standard bivariate normal distribution, where $\phi_2(x_1, x_2; t)$ is the probability density function of the standard bivariate normal distribution with correlation $t$.

By (2), it is seen that the population version of Kendall's tau $\tau_{jk} = E(\widehat{\tau}_{jk})$ is

$$\begin{aligned} \tau_{jk} &= 2E(X_{ij}X_{ik}) - 2E(X_{ij})E(X_{ik}) \\ &= 2\mathbb{P}\{f_j(Z_{ij}) > \Delta_j, f_k(Z_{ik}) > \Delta_k\} - 2\mathbb{P}\{f_j(Z_{ij}) > \Delta_j\} \cdot \mathbb{P}\{f_k(Z_{ik}) > \Delta_k) \\ &= 2\Big\{\Phi_2(\Delta_j, \Delta_k, \Sigma_{jk}) - \Phi(\Delta_j)\Phi(\Delta_k)\Big\}. \end{aligned} \quad (3)$$

Denote by

$$F(t; \Delta_j, \Delta_k) = 2\Big\{\Phi_2(\Delta_j, \Delta_k, t) - \Phi(\Delta_j)\Phi(\Delta_k)\Big\}.$$

For fixed $\Delta_j$ and $\Delta_k$, $F(t; \Delta_j, \Delta_k)$ serves as a bridge function that connects the latent correlation $\Sigma_{jk}$ to the population version of Kendall's tau $\tau_{jk}$. For instance, when $\Delta_j = \Delta_k = 0$, by the Sheppard's theorem (Sheppard, 1899), we can show that $F(t; 0, 0) = \frac{1}{\pi}\sin^{-1} t$, that is $\tau_{jk} = \frac{1}{\pi}\sin^{-1}\Sigma_{jk}$. By the substitution method, $\Sigma_{jk}$ can be consistently estimated by $\sin(\pi\widehat{\tau}_{jk})$. To find a consistent estimator of $\mathbf{\Sigma}$ in the general setting, we need the following lemma, whose proof can be found in the Appendix.

**Lemma 3.1.** *For any fixed $\Delta_j$ and $\Delta_k$, $F(t; \Delta_j, \Delta_k)$ is a strictly monotonic increasing function on $t \in (-1, 1)$.*

Let $\tau = F(t; \Delta_j, \Delta_k)$. By Lemma 3.1, for fixed $\Delta_j$ and $\Delta_k$, $F(t; \Delta_j, \Delta_k)$ is invertible, and we denote the inverse function by $F^{-1}(\tau; \Delta_j, \Delta_k)$. Given $\Delta_j$ and $\Delta_k$, one can estimate $\Sigma_{jk}$ by $F^{-1}(\widehat{\tau}_{jk}; \Delta_j, \Delta_k)$. However, in practice, the cutoff values $\Delta_j$ and $\Delta_k$ are unknown. They can be estimated from the equation

$$E(X_{ij}) = 1 - \Phi(\Delta_j), \quad (4)$$

as noted before. Namely, $\Delta_j$ can be estimated by $\widehat{\Delta}_j = \Phi^{-1}(1 - \bar{X}_j)$, where $\bar{X}_j = \sum_{i=1}^n X_{ij}/n$.

**Definition 3.2.** Define the Kendall's tau correlation matrix estimator $\widehat{\mathbf{R}} = [\widehat{R}_{jk}]$ as a $d \times d$ matrix with element entry given by

$$\widehat{R}_{jk} = \begin{cases} F^{-1}(\widehat{\tau}_{jk}; \widehat{\Delta}_j, \widehat{\Delta}_k), & j \neq k, \\ 1, & j = k. \end{cases} \quad (5)$$

To study the theoretical properties of $\widehat{\mathbf{R}}$, we assume the following regularity conditions.

(A1) There exists a constant $\delta > 0$ such that $|\Sigma_{jk}| \leq 1 - \delta$, for any $j \neq k = 1, ..., d$.

(A2) There exists a constant $M$ such that $|\Delta_j| \leq M$, for any $j = 1, ..., d$.

Conditions (A1) and (A2) are mainly adopted for technical considerations and they impose little restriction in practice. Specifically, Condition (A1) rules out the singular case that $Z_{ij}$ and $Z_{ik}$ are perfectly collinear. Condition (A2) is used to control the variation of $F^{-1}(\tau; \Delta_j, \Delta_k)$ with respect to $(\tau; \Delta_j, \Delta_k)$. Under these conditions, we obtain the following lemma, which is the key to establish the theoretical properties of $\widehat{\mathbf{R}}$.

**Lemma 3.2.** *Under Conditions (A1) and (A2), $F^{-1}(\tau; \Delta_j, \Delta_k)$ is Lipschitz in $\tau$ uniformly over $\Delta_k$ and $\Delta_j$, i.e., there exists a Lipschitz constant $L_2$ independent of $(\Delta_j, \Delta_k)$, such that*

$$|F^{-1}(\tau_1; \Delta_j, \Delta_k) - F^{-1}(\tau_2; \Delta_j, \Delta_k)| \leq L_2|\tau_1 - \tau_2|. \tag{6}$$

The following theorem establishes the $O_p(\sqrt{\log d/n})$ convergence rate of $\widehat{R}_{jk} - \Sigma_{jk}$ uniformly over $j, k$.

**Theorem 3.1.** *Under assumptions (A1) and (A2), for any $t > 0$ we have,*

$$\mathbb{P}\left(||\widehat{\mathbf{R}} - \mathbf{\Sigma}||_{\max} > t\right) \leq 2d^2 \exp\left(-\frac{nt^2}{8L_2^2}\right) + 4d^2 \exp\left(-\frac{nt^2\pi}{16^2 L_1^2 L_2^2}\right) + 4d^2 \exp\left(-\frac{M^2 n}{2L_1^2}\right), \tag{7}$$

*where $L_1$ is a positive constant given in Lemma A.1 in Appendix A.3 and $L_2$ is a positive constant given in Lemma 3.2. That is, for some constant $C$, $\sup_{j,k} |\widehat{R}_{jk} - \Sigma_{jk}| \leq C\sqrt{\log d/n}$ with probability greater than $1 - d^{-1}$.*

# 4 Latent Graph Structure Learning for Binary Data

The structure of the latent graph is characterized by the sparsity pattern of the inverse correlation matrix $\mathbf{\Omega}$. It is seen that the likelihood based method such as that in Han and Pan (2012) is not feasible, because of the complicated form of the likelihood and score functions. In this section, we propose a simple method for modifying the estimators under the Gaussian graphical model to estimate $\mathbf{\Omega}$. The basic idea is to replace the sample covariance matrix in Gaussian graphical models by the rank based covariance matrix estimator. For concreteness, we confine our attention to the graphical Lasso estimator (Friedman et al., 2008), CLIME estimator (Cai et al., 2011), and adaptive graphical Lasso estimator (Fan et al., 2009, 2012) and indicate that similar procedures can be also applied to other existing Gaussian graph estimators with the sample covariance matrix as the input.

## 4.1 Graphical Lasso Estimator

As illustrated by Friedman et al. (2008); d'Aspremont et al. (2008); Banerjee et al. (2008), the graphical Lasso estimator of $\mathbf{\Omega}$ in the Gaussian model is

$$\bar{\mathbf{\Omega}} = \underset{\mathbf{\Omega} \succeq 0}{\arg\min} \left\{ \mathrm{Tr}(\widehat{\mathbf{\Sigma}}\mathbf{\Omega}) - \log|\mathbf{\Omega}| + \lambda \sum_{j \neq k} |\Omega_{jk}| \right\}, \tag{8}$$

where $\lambda$ is a regularization parameter and $\widehat{\mathbf{\Sigma}} = \sum_{i=1}^n \boldsymbol{Z}_i \boldsymbol{Z}_i^T/n$ is the sample covariance matrix of Gaussian random variables $\boldsymbol{Z}_1, ...., \boldsymbol{Z}_n$. However, in our latent Gaussian copula model, $\bar{\mathbf{\Omega}}$ is not computable, because the latent variable $\boldsymbol{Z}_i$ is not observed. Our proposal is to replace the sample covariance matrix $\widehat{\boldsymbol{\Sigma}}$ with the rank based estimator $\widehat{\mathbf{R}}$ in (5), and solve the following optimization problem,

$$\widetilde{\boldsymbol{\Omega}} = \mathop{\arg\min}_{\boldsymbol{\Omega}\succeq 0} \Big\{ \mathrm{Tr}(\widehat{\mathbf{R}}\boldsymbol{\Omega}) - \log|\boldsymbol{\Omega}| + \lambda \sum_{j\neq k} |\Omega_{jk}| \Big\}. \tag{9}$$

However, one potential issue with the rank-based estimator is that $\widehat{\mathbf{R}}$ may not be positive semidefinite. Since we do not penalize the diagonal elements of $\boldsymbol{\Omega}$, the diagonal elements of $\widetilde{\boldsymbol{\Omega}}$ can diverge to infinity. Even though the optimization problem in (9) remains convex, the coordinate descent algorithm in Friedman et al. (2008) may not converge. To further regularize the estimator, we can project $\widehat{\mathbf{R}}$ into the cone of positive semidefinite matrices, i.e.,

$$\widehat{\mathbf{R}}_p = \mathop{\arg\min}_{\boldsymbol{R}\succeq 0} ||\widehat{\mathbf{R}} - \boldsymbol{R}||_{\max}. \tag{10}$$

The smoothed approximation method in Nesterov (2005) can be used to calculate $\widehat{\mathbf{R}}_p$; see also Liu et al. (2012). With $\widehat{\mathbf{R}}$ in (9) replaced by $\widehat{\mathbf{R}}_p$, we obtain the graphical Lasso estimator

$$\widehat{\boldsymbol{\Omega}}_G = \mathop{\arg\min}_{\boldsymbol{\Omega}\succeq 0} \Big\{ \mathrm{Tr}(\widehat{\mathbf{R}}_p\boldsymbol{\Omega}) - \log|\boldsymbol{\Omega}| + \lambda \sum_{j\neq k} |\Omega_{jk}| \Big\}. \tag{11}$$

Due to the positive semidefiniteness of $\widehat{\mathbf{R}}_p$, the convergence properties of the computational algorithms such as the coordinate descent for solving (11) are identical to those for solving the Gaussian graphical Lasso in (8). We refer to Rothman et al. (2008); Friedman et al. (2008) for details.

### 4.2 CLIME Estimator

Following the similar procedures in Cai et al. (2011), we can replace $\widehat{\boldsymbol{\Sigma}}$ with $\widehat{\mathbf{R}}$ to construct the CLIME estimator of $\boldsymbol{\Omega}$. Consider the following intermediate estimator:

$$\widehat{\boldsymbol{\Omega}}_1 = \mathop{\arg\min}_{\boldsymbol{\Omega}\in\mathbb{R}^{d\times d}} ||\boldsymbol{\Omega}||_1, \quad \text{s.t.} \quad ||\widehat{\mathbf{R}}\boldsymbol{\Omega} - \mathbf{I}_d||_{\max} \le \lambda, \tag{12}$$

where $\lambda$ is a regularization parameter. Let $\widehat{\boldsymbol{\Omega}}_1 = (\widehat{\omega}^1_{ij})$. To ensure that the estimator of $\Omega$ is symmetric, we take either $\widehat{\omega}^1_{ij}$ or $\widehat{\omega}^1_{ji}$ depending on whose magnitude is smaller. After this symmetrization step, we obtain the CLIME estimator $\widehat{\boldsymbol{\Omega}}_C = (\widehat{\omega}_{Cij})$, where

$$\widehat{\omega}_{Cij} = \widehat{\omega}_{Cji} = \widehat{\omega}^1_{ij} I(|\widehat{\omega}^1_{ij}| \le \widehat{\omega}^1_{ji}) + \widehat{\omega}^1_{ji} I(|\widehat{\omega}^1_{ij}| > \widehat{\omega}^1_{ji}).$$

The same computational procedure in Cai et al. (2011) can be used to solve the convex program (12). We refer to Cai et al. (2011) for details.

### 4.3 Adaptive Graphical Lasso Estimator

It has been well known that, due to the bias of the $\ell_1$ penalty, the graphical Lasso estimator such as $\bar{\boldsymbol{\Omega}}$ in (8) requires strong model assumptions such as mutual incoherence or irrepresentable conditions to achieve model selection consistency (Ravikumar et al., 2011). To reduce the bias of the $\ell_1$ penalty, maximum penalized likelihood estimation with nonconvex penalty functions is considered by Lam and Fan (2009) for the Gaussian graphical model. To estimate the precision matrix in our latent Gaussian copula model, we can similarly minimize the following nonconvex function,

$$\mathrm{Tr}(\widehat{\mathbf{R}}_p\mathbf{\Omega}) - \log|\mathbf{\Omega}| + \sum_{j\neq k} p_\lambda(|\Omega_{jk}|), \tag{13}$$

where $p_\lambda(\theta)$ is a nonconvex penalty function with regularization parameter $\lambda$. One such example is the SCAD penalty (Fan and Li, 2001) given by

$$p_\lambda(\theta) = \int_0^{|\theta|}\left\{\lambda I(z\leq\lambda) + \frac{(a\lambda - z)_+}{a-1}I(z>\lambda)\right\}dz,$$

for some $a > 2$. However, minimizing (13) is challenging due to the nonconvexity of the penalty function and the computed solution may heavily depend on the specific algorithm for minimizing (13). To address this issue, a computationally efficient local linear approximation (LLA) method is proposed by Zou and Li (2008); Fan et al. (2012), which was inspired by the local quadratic approximation method of Fan and Li (2001). The key idea of the LLA method is to use a linear function to approximate the nonconvex penalty $p_\lambda(\theta)$ and turn the nonconvex optimization problem (13) into a sequence of weighted $\ell_1$ penalization problem. In our context, we can simply solve the following convex problem

$$\widehat{\mathbf{\Omega}}_A = \operatorname*{arg\,min}_{\mathbf{\Omega}\succeq 0}\left\{\mathrm{Tr}(\widehat{\mathbf{R}}_p\mathbf{\Omega}) - \log|\mathbf{\Omega}| + \sum_{j\neq k} p'_\lambda(|\widehat{\Omega}^0_{jk}|)|\Omega_{jk}|\right\}, \tag{14}$$

where $p'_\lambda(\theta)$ is the derivative of $p_\lambda(\theta)$ with respect to $\theta$ and $\widehat{\mathbf{\Omega}}^0 = (\widehat{\Omega}^0_{jk})$ is an initial estimator of $\mathbf{\Omega}$ which can be taken as the graphical Lasso estimator $\widehat{\mathbf{\Omega}}_G$ or the CLIME estimator $\widehat{\mathbf{\Omega}}_C$. The penalty $p'_\lambda(|\widehat{\Omega}^0_{jk}|)|\Omega_{jk}|$ in (14) adaptively shrinks the estimator of $\Omega_{jk}$, and hence we call $\widehat{\mathbf{\Omega}}_A$ as the adaptive graphical Lasso estimator.

### 4.4 Theoretical Properties

Recall that in Theorem 3.1, we establish the error bound of $\widehat{\mathbf{R}}$ under the $\ell_\infty$ norm. The following theorem shows that similar error bound holds for the projected estimator $\widehat{\mathbf{R}}_p$ in (10).

**Theorem 4.1.** *Under assumptions (A1) and (A2), for any $t > 0$ we have,*

$$\mathbb{P}\left(||\widehat{\mathbf{R}}_p - \mathbf{\Sigma}||_{\max} > 2t\right) \leq 2d^2\exp\left(-\frac{nt^2}{8L_2^2}\right) + 4d^2\exp\left(-\frac{nt^2\pi}{16^2L_1^2L_2^2}\right) + 4d^2\exp\left(-\frac{M^2n}{2L_1^2}\right), \tag{15}$$

*where $L_1$ is a positive constant given in Lemma A.1 in Appendix A.3 and $L_2$ is a positive constant given in Lemma 3.2. That is, for some constant $C$ independent of $d$ and $n$, $||\widehat{\mathbf{R}}_p - \mathbf{\Sigma}||_{\max} \leq C\sqrt{\log d/n}$ with probability greater than $1 - d^{-1}$.*

By Theorems 3.1 and 4.1, we find that the error bounds of the rank based estimators $\widehat{\mathbf{R}}$ and $\widehat{\mathbf{R}}_p$ are similar to that of the sample covariance matrix. Therefore, the estimators $\widehat{\mathbf{\Omega}}_G$, $\widehat{\mathbf{\Omega}}_C$ and $\widehat{\mathbf{\Omega}}_A$ enjoy the same theoretical properties as those established by Raskutti et al. (2008), Cai et al.

(2011) and Fan et al. (2012) under the Gaussian graphical model. For simplicity of presentation, we mainly confine our attention to $\widehat{\mathbf{\Omega}}_C$ and $\widehat{\mathbf{\Omega}}_A$.

Similar to Cai et al. (2011), for $0 \leq q < 1$, $K > 0$ and $s_0 > 0$, consider the class of matrices

$$\mathcal{U}(q, s_0, K) = \left\{ \mathbf{\Omega} : \mathbf{\Omega} \succ 0, ||\mathbf{\Omega}||_{L_1} \leq K, \max_{1 \leq i \leq d} \sum_{j=1}^{d} |\omega_{ij}|^q \leq s_0 \right\},$$

where $\mathbf{\Omega} = (\omega_{ij})$. For a matrix $A = (a_{ij})$, denote $\mathcal{M}(A) = \{\text{sign}(a_{ij}), 1 \leq i, j \leq d\}$ and $\mathcal{S} = \{(i, j) : \Omega_{ij} \neq 0\}$ to be the support set of $\mathbf{\Omega}$. We can show the following theorem on the rate of convergence and graph recovery consistency of $\widehat{\mathbf{\Omega}}_C$.

**Theorem 4.2.** *Assume that Conditions (A1) and (A2) hold, $\mathbf{\Omega} \in \mathcal{U}(q, s_0, K)$, and $\lambda = C_0 K \sqrt{\log d / n}$, for some constant $C_0 > 0$. Then with probability greater than $1 - d^{-1}$, we have*

*(1) (Rate of convergence)*

$$||\widehat{\mathbf{\Omega}}_C - \mathbf{\Omega}||_{\max} \leq 4 C_0 K^2 \sqrt{\frac{\log d}{n}}$$

$$||\widehat{\mathbf{\Omega}}_C - \mathbf{\Omega}||_2 \leq C_1 K^{2-2q} s_0 \left( \frac{\log d}{n} \right)^{(1-q)/2}$$

$$\frac{1}{d} ||\widehat{\mathbf{\Omega}}_C - \mathbf{\Omega}||_F \leq 4 C_1 K^{4-2q} s_0 \left( \frac{\log d}{n} \right)^{(1-q)/2},$$

*where $C_1$ is a constant only depending on $C_0$ and $q$.*

*(2) (Sign consistency) Let $\widetilde{\mathbf{\Omega}}_C = (\widetilde{\omega}_{Cij})$ with $\widetilde{\omega}_{Cij} = \widehat{\omega}_{Cij} I(|\widehat{\omega}_{Cij}| \geq \tau)$, where $\tau > 4K\lambda$. If $\min_{(i,j) \in \mathcal{S}} |\omega_{ij}| > 2\tau$, then with probability tending to one, $\mathcal{M}(\widetilde{\Omega}_C) = \mathcal{M}(\Omega)$.*

By Theorem 4.2, the proposed CLIME estimator achieves the same rate of convergence for both matrix estimation and graph recovery, as those if the latent variables $\boldsymbol{Z}_1, ..., \boldsymbol{Z}_n$ were observed.

Fan et al. (2012) established the strong oracle property of the LLA method under the Gaussian graphical model. Their theoretical development hinges on the concentration results of the sample covariance matrix, which are similar to (7) and (15). Define the oracle estimator by

$$\widehat{\mathbf{\Omega}}^{(orc)} = \underset{\mathbf{\Omega} \succeq 0, \mathbf{\Omega}_{\mathcal{S}} = 0}{\arg\min} \left\{ \text{Tr}(\widehat{\mathbf{R}}_p \mathbf{\Omega}) - \log |\mathbf{\Omega}| \right\},$$

which is the estimator of $\mathbf{\Omega}$ under the assumption that the support set $\mathcal{S}$ is known. Under the same conditions as those in Fan et al. (2012), the adaptive graphical Lasso estimator $\widehat{\mathbf{\Omega}}_A$ satisfies the strong oracle property, i.e., $\widehat{\mathbf{\Omega}}_A = \widehat{\mathbf{\Omega}}^{(orc)}$ with high probability. A nonasymptotic result can be established following the same line as in Fan et al. (2012), and is omitted.

## 5 Latent Principal Component Analysis for Binary Data

In principal component analysis, the important patterns in the data are represented by the leading eigenvector of the latent correlation matrix $\mathbf{\Sigma}$. In this section, we consider estimation of the leading eigenvectors of $\mathbf{\Sigma}$ and study the theoretical properties.

## 5.1 Eigenvector Estimation via Two-stage Method

By definition, the leading eigenvector of $\mathbf{\Sigma}$, denoted by $\boldsymbol{v}_1 \in \mathbb{R}^d$, corresponds to the maximizer of $\boldsymbol{v}^T \Sigma \boldsymbol{v}$ subject to the constraint that $||\boldsymbol{v}||_2 = 1$. Given the rank based estimator $\widehat{\mathbf{R}}$, one can estimate the leading eigenvector of $\mathbf{\Sigma}$ by the following sparse PCA

$$\widehat{\boldsymbol{v}}_1(k) = \arg\max_{\boldsymbol{v} \in \mathbb{R}^d} \boldsymbol{v}^T \widehat{\mathbf{R}} \boldsymbol{v}, \quad \text{s.t.} \quad \boldsymbol{v} \in \mathbb{S}^{d-1} \cap \mathbb{B}_0(k), \tag{16}$$

where $\mathbb{B}_0(k) = \{\boldsymbol{v} \in \mathbb{R}^d : ||\boldsymbol{v}||_0 \leq k\}$ and $k$ is a tuning parameter that governs the sparsity level of the vector $\boldsymbol{v}$. If the maximizers in (16) are not unique, we allow $\widehat{\boldsymbol{v}}_1(k)$ to be any one of them. In the following section, we can show that any maximizer in (16) has provable theoretical guarantees. Due to the cardinality constraint, the calculation of $\widehat{v}_1(k)$ is NP-hard and therefore intractable. To circumvent this issue, d'Aspremont et al. (2004); Vu et al. (2013) proposed an alternative estimator resulted from a semidefinite program, which can be computed at the polynomial time. Note that (16) can be reformulated as a penalization problem

$$\arg\max_{\boldsymbol{v} \in \mathbb{R}^d} \langle \widehat{\mathbf{R}}, \boldsymbol{v}\boldsymbol{v}^T \rangle - \lambda ||\boldsymbol{v}||_0^2, \quad \text{s.t.} \quad \boldsymbol{v} \in \mathbb{S}^{d-1}, \tag{17}$$

which is still computationally intractable. For any $\boldsymbol{v} \in \mathbb{S}^{d-1}$ and $\boldsymbol{W} = \boldsymbol{v}\boldsymbol{v}^T$, we have $\mathrm{Tr}(\boldsymbol{W}) = 1$, and a simple inequality yields,

$$||\boldsymbol{W}||_{1,1} = \sum_{i,j} |v_i v_j| \leq ||\boldsymbol{v}||_0^2.$$

Thus, the convex relaxation of (17) is given by

$$\widetilde{\boldsymbol{W}} = \arg\max_{\boldsymbol{W} \succeq 0} \langle \widehat{\mathbf{R}}, \boldsymbol{W} \rangle - \lambda ||\boldsymbol{W}||_{1,1}, \quad \text{s.t.} \quad \mathrm{Tr}(\boldsymbol{W}) = 1, \tag{18}$$

where $\lambda$ is a regularization parameter. Since $\widetilde{\boldsymbol{W}}$ is at least with rank 1, we can take the leading eigenvector of $\widetilde{\boldsymbol{W}}$ as the estimator of $\boldsymbol{v}_1$, denoted by $\widetilde{\boldsymbol{v}}_1$. It has been shown that such an estimator $\widetilde{\boldsymbol{v}}_1$ is unique. The semidefinite program (18) can be solved efficiently using alternating direction method of multipliers (ADMM) algorithm (Boyd et al., 2011). We refer to Vu et al. (2013) for more details.

While $\widetilde{\boldsymbol{v}}_1$ is computationally simple, it is not minimax optimal (Vu et al., 2013). A potential option to construct the optimal estimator is to use the following two-stage procedure. In the first stage, the ADMM algorithm is used to maximize (18) and obtain the estimator $\widetilde{\boldsymbol{v}}_1$. In the second stage, we use the truncated power method (Yuan and Zhang, 2013) with $\widetilde{\boldsymbol{v}}_1$ as an initial estimator to construct a new estimator $\widetilde{\boldsymbol{v}}_1(k)$ which has $k$ nonzero components. The truncated power algorithm is described as follows.

(i) Given the rank based estimator $\widehat{\mathbf{R}}$, the initial vector $\boldsymbol{v}^{(0)} = \widetilde{\boldsymbol{v}}_1$, and the tuning parameter $k$, set $t = 0$.

(ii) Set $t \leftarrow t + 1$. Compute $\mathbf{x}_t = \widehat{\mathbf{R}}\boldsymbol{v}^{(t-1)}$. If $||\mathbf{x}_t||_0 \leq k$, then take

$$\boldsymbol{v}^{(t)} = \mathbf{x}_t / ||\mathbf{x}_t||_2.$$

If $||\mathbf{x}_t||_0 > k$, then take

$$\boldsymbol{v}^{(t)} = \text{TRC}(\mathbf{x}_t, A_t)/||\,\text{TRC}(\mathbf{x}_t, A_t)||_2,$$

where $A_t$ is the indices of the largest $k$ components of $\mathbf{x}_t$ in absolute values, and

$$\text{TRC}(\mathbf{x}_t, A_t) := \big(x_{t1} \cdot I(1 \in A_t),\ \ \ldots,\ x_{td} \cdot I(d \in A_t)\big)^T.$$

(iii) Repeat (ii) until convergence. Then set $\widetilde{\boldsymbol{v}}_1(k) = \boldsymbol{v}^{(\infty)}$.

The estimator $\widetilde{\boldsymbol{v}}_1$ in the first stage, although suboptimal, is adequate to initialize the algorithm. In step (ii), an iterative procedure is used to project $\mathbf{x}_t$ to $\mathbb{S}^{d-1} \cap \mathbb{B}_0(k)$. Specifically, when $||\mathbf{x}_t||_0 \le k$ holds, we simply rescale $\mathbf{x}_t$ such that $||\boldsymbol{v}^{(t)}||_2 = 1$. When $||\mathbf{x}_t||_0 > k$, we keep the largest $k$ components of $\mathbf{x}_t$ in absolute values, truncate the remaining components to 0, and rescale $\text{TRC}(\mathbf{x}_t, A_t)$ to that with $||\boldsymbol{v}^{(t)}||_2 = 1$. Finally, we repeat the procedure until certain stopping rule is met such as $||\boldsymbol{v}^{(t)} - \boldsymbol{v}^{(t-1)}||_2 \le \epsilon$, where $\epsilon$ is a small positive number. The convergence property of the algorithm is established by Yuan and Zhang (2013). Note that the algorithm results in the estimator $\widetilde{\boldsymbol{v}}_1(k)$ that is typically different from the global maximizer $\widehat{\boldsymbol{v}}_1(k)$.

## 5.2 Theoretical Properties

In this section, we establish the theoretical properties of $\widetilde{\boldsymbol{v}}_1(k)$ and $\widehat{\boldsymbol{v}}_1(k)$. To quantify the angle between two vectors $\boldsymbol{u}_1 \in \mathbb{S}^{d-1}$ and $\boldsymbol{u}_2 \in \mathbb{S}^{d-1}$, we define $|\sin\angle(\boldsymbol{u}_1, \boldsymbol{u}_2)| = \sqrt{1 - (\boldsymbol{u}_1^T\boldsymbol{u}_2)^2}$. The follow theorem shows the upper bounds for the angles between $\widehat{\boldsymbol{v}}_1(k)$ and $\boldsymbol{v}_1$, and between $\widetilde{\boldsymbol{v}}_1(k)$ and $\boldsymbol{v}_1$.

**Theorem 5.1.** *Assume that Conditions (A1) and (A2) hold, and the leading eigenvector $\boldsymbol{v}_1$ satisfies $||\boldsymbol{v}_1||_0 = k_0$. For $k \ge k_0$, with probability greater than $1 - d^{-1}$, we have*

$$|\sin\angle(\widehat{\boldsymbol{v}}_1(k), \boldsymbol{v}_1)| \le \frac{2C}{\lambda_1 - \lambda_2} \cdot k\sqrt{\frac{\log d}{n}}, \tag{19}$$

*where $\lambda_j = \Lambda_j(\boldsymbol{\Sigma})$ for $j = 1, 2$ and $C$ is given in Theorem 3.1. In addition, under the conditions in Theorem 1 in Yuan and Zhang (2013), we have, with probability greater than $1 - d^{-1}$,*

$$|\sin\angle(\widetilde{\boldsymbol{v}}_1(k), \boldsymbol{v}_1)| \le C' \cdot (k_0 + 2k)\sqrt{\frac{\log d}{n}}, \tag{20}$$

*for some constant $C'$ independent of $(n, d, k_0)$.*

By Theorem 5.1, we find that $|\sin\angle(\widehat{\boldsymbol{v}}_1(k), \boldsymbol{v}_1)| = O_p(\sqrt{\log d/n})$ and $|\sin\angle(\widetilde{\boldsymbol{v}}_1(k), \boldsymbol{v}_1)| = O_p(\sqrt{\log d/n})$ for fixed $k_0$, $k$, $\lambda_1$ and $\lambda_2$, which agrees with the minimax optimal rate shown in Vu and Lei (2012) if the latent variables $\boldsymbol{Z}_1, ..., \boldsymbol{Z}_n$ were observed. However, if we allow $k_0$ to increase with $n$, the statistical rate in Theorem 5.1 is suboptimal. Recently, Wang and Liu (2014) conducted a more refined analysis for the truncated power method and showed that under additional conditions, $|\sin\angle(\widetilde{\boldsymbol{v}}_1(k), \boldsymbol{v}_1)| = O_p(\sqrt{k\log d/n})$, which improves (20) and achieves the optimal scaling for $k$. The detailed results will be reported in a technical report by Wang and Liu (2014).

Let $V_1 = \mathrm{supp}(\boldsymbol{v}_1)$, $\widehat{V}_1(k) = \mathrm{supp}(\widehat{\boldsymbol{v}}_1(k))$, and $\widetilde{V}_1(k) = \mathrm{supp}(\widetilde{\boldsymbol{v}}_1(k))$ denote the support sets for $\boldsymbol{v}_1$, $\widehat{\boldsymbol{v}}_1(k)$ and $\widetilde{\boldsymbol{v}}_1(k)$. The follow theorem establishes the sure screening property of $\widehat{\boldsymbol{v}}_1(k)$ and $\widetilde{\boldsymbol{v}}_1(k)$, following the terminology of Fan and Lv (2008).

**Theorem 5.2.** *Assume that Conditions (A1) and (A2) hold and the leading eigenvector $\boldsymbol{v}_1$ satisfies $||\boldsymbol{v}_1||_0 = k_0$ with $k \geq k_0$. For $C$ given in Theorem 3.1, if $\min_{j\in V_1} |v_{1j}| \geq \frac{2\sqrt{2}C}{\lambda_1-\lambda_2} \cdot k\sqrt{\frac{\log d}{n}}$, then we have*

$$\mathbb{P}(V_1 \subset \widehat{V}_1(k)) \geq 1 - d^{-1}.$$

*If $\min_{j\in V_1} |v_{1j}| \geq \sqrt{2}C' \cdot (k_0 + 2k)\sqrt{\frac{\log d}{n}}$, where $C'$ is given in Theorem 5.1, then we have*

$$\mathbb{P}(V_1 \subset \widetilde{V}_1(k)) \geq 1 - d^{-1}.$$

This theorem suggests that under similar assumptions on the minimum signal strength of $\boldsymbol{v}_1$ to those for the Gaussian data or the Gaussian copula data (Han and Liu, 2012), feature selection consistency of $\widehat{\boldsymbol{v}}_1(k)$ and $\widetilde{\boldsymbol{v}}_1(k)$ can be achieved. The proof of Theorem 5.2 hinges on the convergence rate in Theorem 5.1 and the minimal signal strength conditions. Once an improved rate for $\widetilde{\boldsymbol{v}}_1(k)$ is obtained such as that in Wang and Liu (2014), the minimal signal strength condition can be relaxed.

# 6 Latent Gaussian Copula Model for Mixed Data

## 6.1 A Model for Mixed Data

We start with the following definition of the latent Gaussian copula model for mixed data.

**Definition 6.1 (Latent Gaussian copula model for mixed data).** Assume that $\mathbf{X} = (\mathbf{X}_1, \mathbf{X}_2)$, where $\mathbf{X}_1$ represent the $d_1$-dimensional continuous variables and $\mathbf{X}_2$ represent the $d_2$-dimensional binary variables. The random vector $\mathbf{X}$ satisfies the latent Gaussian copula model, if there exists a $d_2$-dimensional random vector $\boldsymbol{Z}_2 = (Z_{d_1+1}, ..., Z_d)^T$ such that $\boldsymbol{Z} := (\mathbf{X}_1, \boldsymbol{Z}_2) \sim \mathrm{NPN}(0, \boldsymbol{\Sigma}, f)$ and

$$X_j = I(Z_j > C_j) \ \text{ for all } j = d_1 + 1, \ldots, d,$$

where $\boldsymbol{C} = (C_{d_1+1}, ..., C_d)$ is a vector of constants. Then we denote $\mathbf{X} \sim \mathrm{LNPN}(0, \boldsymbol{\Sigma}, f, \boldsymbol{C})$, and call $\boldsymbol{\Sigma}$ the latent correlation matrix. When $\boldsymbol{Z} \sim \mathrm{N}(0, \boldsymbol{\Sigma})$, we say $\mathbf{X}$ satisfies the latent Gaussian model $\mathrm{LN}(0, \boldsymbol{\Sigma}, \boldsymbol{C})$.

In the latent Gaussian copula model, the binary components $\mathbf{X}_2$ are generated by a latent continuous random vector $\boldsymbol{Z}_2$ truncated at $\boldsymbol{C}$, and combining with the continuous components $\mathbf{X}_1$, $\boldsymbol{Z} = (\mathbf{X}_1, \boldsymbol{Z}_2)$ satisfies the Gaussian copula model. Following the similar arguments in Section 3, for the binary components, only $\Delta_j = f_j(C_j)$, $j = d_1 + 1, ..., d$, are identifiable. For the continuous components, the marginal transformations $f_j(\cdot)$, $j = 1, ..., d_1$, are identifiable. Hence, the family of latent Gaussian copula model is strictly larger than the latent Gaussian model for mixed data. Let $\boldsymbol{\Omega} = \boldsymbol{\Sigma}^{-1}$ be the latent precision matrix. Similar to Section 3, the zero patterns of $\Omega$ characterize the conditional independence structure among the latent variables $\boldsymbol{Z}$.

## 6.2 Rank Based Latent Correlation Matrix Estimator

Assume that we observe $n$ independent vector-valued data $\mathbf{X}_1, \ldots, \mathbf{X}_n \sim \text{LNPN}(0, \boldsymbol{\Sigma}, f, \boldsymbol{C})$. As shown in Section 3, the rank based correlation matrix estimator $\widehat{R}_{jk}$ given by (5) is a consistent estimator of $\Sigma_{jk}$ for discrete $X_{ij}$ and $X_{ik}$. To study the theoretical properties of $\widehat{R}_{jk}$, similar to (A1) and (A2), we assume the following conditions.

(A1') There exists a constant $\delta > 0$ such that $|\Sigma_{jk}| < 1 - \delta$, for any $j \neq k = d_1 + 1, ..., d$.

(A2') There exists a constant $M$ such that $|\Delta_j| \leq M$, for any $j = d_1 + 1, ..., d$.

For the discrete components, we have the following result, which is identical to Theorem 3.1 with a slight change of notation.

**Lemma 6.1.** *Under assumptions (A1') and (A2'), with probability greater than $1 - d_2^{-1}$, we have*

$$\sup_{d_1+1 \leq j,k \leq d} |\widehat{R}_{jk} - \Sigma_{jk}| \leq C\sqrt{\frac{\log d_2}{n}},$$

*where $C$ is a constant independent of $(n, d_2)$.*

Given the Kendall's tau $\widehat{\tau}_{jk}$ in (2), when $X_{ij}$ and $X_{ik}$ are both continuous, Liu et al. (2012) showed that the rank based estimator is

$$\widehat{R}_{jk} = \begin{cases} \sin\left(\frac{\pi}{2}\widehat{\tau}_{jk}\right), & j \neq k, \\ 1, & j = k. \end{cases} \tag{21}$$

Analogous to Lemma 6.1, for the continuous components, the following lemma provides the upper bound for $|\widehat{R}_{jk} - \Sigma_{jk}|$.

**Lemma 6.2.** *For $n > 1$, with probability greater than $1 - d_1^{-1}$, we have*

$$\sup_{1 \leq j,k \leq d_1} |\widehat{R}_{jk} - \Sigma_{jk}| \leq 2.45\pi\sqrt{\frac{\log d_1}{n}}.$$

This result follows directly from Theorem 4.2 in Liu et al. (2012). Now we consider the mixed case. Without loss of generality, we assume that $X_{ij}$ is binary and $X_{ik}$ is continuous. In this case, the Kendall's tau is given by

$$\widehat{\tau}_{jk} = \frac{2}{n(n-1)} \sum_{1 \leq i < i' \leq n} (X_{ij} - X_{i'j}) \text{sign}(X_{ik} - X_{i'k}).$$

The following lemma establishes the bridge function that connects the population version of Kendall's tau to $\boldsymbol{\Sigma}$ for the mixed data.

**Lemma 6.3.** *The the population version of Kendall's tau correlation $\tau_{jk} = E(\widehat{\tau}_{jk})$ is given by $\tau_{jk} = H(\Sigma_{jk}; \Delta_j)$, where*

$$H(t; \Delta_j) = 4\Phi_2(\Delta_j, 0, t/\sqrt{2}) - 2\Phi(\Delta_j). \tag{22}$$

*Moreover, for fixed $\Delta_j$, $H(t; \Delta_j)$ is an invertible function of $t$. In particular, when $\Delta_j = 0$, we have $H(t, 0) = \frac{2}{\pi}\sin^{-1}(t/\sqrt{2})$, and hence $\Sigma_{jk} = \sqrt{2}\sin(\pi\tau_{jk}/2)$.*

Similar to Section 3, $\Delta_j$ could be estimated by $\widehat{\Delta}_j = \Phi^{-1}(1 - \bar{X}_j)$, where $\bar{X}_j = \sum_{i=1}^n X_{ij}/n$. When $X_{ij}$ is binary and $X_{ik}$ is continuous, the rank based estimator is defined as

$$\widehat{R}_{jk} = \begin{cases} H^{-1}(\widehat{\tau}_{jk}; \widehat{\Delta}_j), & j \neq k, \\ 1, & j = k, \end{cases} \tag{23}$$

where $H^{-1}(\tau, \Delta_j)$ is the inverse function of $H(t, \Delta_j)$ for fixed $\Delta_j$.

**Theorem 6.1.** *Under assumptions (A1') and (A2'), for any $t > 0$ we have,*

$$\begin{aligned}\mathbb{P}\left(\sup_{d_1+1\le j\le d, 1\le k\le d_1} |\widehat{R}_{jk} - \Sigma_{jk}| > t\right) \;\le\; & 2d_1 d_2 \exp\left(-\frac{nt^2}{8L_3^2}\right) + 2d_1 d_2 \exp\left(-\frac{nt^2\pi}{12^2 L_1^2 L_3^2}\right) \\ & + 2d_1 d_2 \exp\left(-\frac{M^2 n}{2L_1^2}\right),\end{aligned}$$

*where $L_1$ is a positive constant given in Lemma A.1 in Appendix A.3 and $L_3$ is a positive constant given in Lemma A.2. That is, for some constant $C$ independent of $(n, d_1, d_2)$,*

$$\sup_{d_1+1\le j\le d, 1\le k\le d_1} |\widehat{R}_{jk} - \Sigma_{jk}| \le C\sqrt{\log d/n}$$

*with probability greater than $1 - d^{-1}$.*

Combining Lemmas 6.1 and 6.2 and Theorem 6.1, we finally obtain the error bound for $\widehat{R}_{jk} - \Sigma_{jk}$ uniformly over $1 \le j, k \le d$.

**Corollary 6.1.** *Under assumptions (A1') and (A2'), with probability greater than $1 - d^{-1}$, we have*

$$\sup_{1\le j,k\le d} |\widehat{R}_{jk} - \Sigma_{jk}| \le C\sqrt{\frac{\log d}{n}},$$

*where $C$ is a constant independent of $(n, d)$.*

Once the rank based estimator $\widehat{\mathbf{R}}$ is obtained, we can adopt the same procedures in Sections 4 and 5 for precision matrix estimation and scale-invariant PCA with the same theoretical guarantees.

# 7 Numerical Results

## 7.1 Simulation Results for Graph Estimation

To evaluate the accuracy of graph estimation, we adopt the similar data generating procedures as in Liu et al. (2012). To generate the inverse correlation matrix $\mathbf{\Omega}$, we set $\Omega_{jj} = 1$, and $\Omega_{jk} = ta_{jk}$, if $j \neq k$, where $t$ is a constant which is chosen to guarantee the positive definiteness of $\mathbf{\Omega}$, and $a_{jk}$ is a Bernoulli random variable with success probability $p_{jk} = (2\pi)^{-1/2}\exp\{||z_j - z_k||_2/(2c_1)\}$, where $z_j = (z_j^{(1)}, z_j^{(2)})$ is independently generated from a bivariate uniform $[0, 1]$ distribution, and $c_1$ is taken as 3. In the simulation studies, we take $d = 50$. This gives a graph with an average of 7.3 edges for each node. Since $\mathbf{\Sigma}$ needs to be a correlation matrix, we rescale $\mathbf{\Omega}$ such that the diagonal elements of $\mathbf{\Sigma}$ are 1.

Assume that the cutoff $C \sim \text{Unif}[-1, 1]$. Consider the following four data generating scenarios.

(a): Simulate data $\mathbf{X} = (X_1, ..., X_d)$, where $X_j = I(Z_j > C_j)$, for all $j = 1, \ldots, d$, and $\boldsymbol{Z} \sim \mathrm{N}(0, \boldsymbol{\Sigma})$.

(b): Simulate data $\mathbf{X} = (X_1, ..., X_d)$, where $X_j = I(Z_j > C_j)$, for all $j = 1, \ldots, d$, and $\boldsymbol{Z} \sim \mathrm{N}(0, \boldsymbol{\Sigma})$, where 5 entries in each $\boldsymbol{Z}$ is randomly sampled and replaced by $-5$ or 5.

(c): Simulate data $\mathbf{X} = (X_1, ..., X_d)$, where $X_j = I(Z_j > C_j)$, for $j = d/2+1, \ldots, d$, $\boldsymbol{Z} \sim \mathrm{N}(0, \boldsymbol{\Sigma})$ and $X_j = Z_j$, for $j = 1, \ldots, d/2$.

(d): Simulate data $\mathbf{X} = (X_1, ..., X_d)$, where $X_j = I(Z_j > C_j)$, for $j = d/2+1, \ldots, d$, $\boldsymbol{Z} \sim \mathrm{NPN}(0, \boldsymbol{\Sigma}, f)$ and $X_j = Z_j$, for $j = 1, \ldots, d/2$, where $f_j(x) = x^3$ for $j = 1, ..., d$.

In Scenarios (a) and (b), the binary data are generated. In particular, Scenario (a) corresponds to the latent Gaussian model and Scenario (b) represents the setting where the binary data can be misclassified due to the outliers of the latent variable. Scenarios (c) and (d) correspond to the mixed data generated from the latent Gaussian model and the latent Gaussian copula model, respectively.

There are 100 replicate simulations, for $n = 50, 100, 200$. For each simulated data set, we apply six estimation methods. That is, the latent graphical Lasso estimator (L-GLASSO) in Section 4.1, the adaptive graphical Lasso estimator (L-GSCAD) in Section 4.3, the approximate sparse maximum likelihood estimator (AMLE) in Banerjee et al. (2008), the naive estimator (Naive) and two graphical Lasso estimators (ZR-GLASSO and ZP-GLASSO) based on the latent variable $\boldsymbol{Z}$. Here, the weight in L-GSCAD is based on the SCAD penalty with $a = 3.7$, the AMLE refers to the graphical Lasso estimator with the modified sample covariance matrix of $\mathbf{X}$ described in Banerjee et al. (2008) as the input, and the naive method is similar to AMLE but with the sample covariance matrix of $\mathbf{X}$ as the input. In ZR-GLASSO and ZP-GLASSO, we assume that $\boldsymbol{Z}$ is observed. In particular, the rank based correlation estimator and the Pearson correlation estimator of $\mathrm{cov}(\boldsymbol{Z})$ is plugged into the graphical Lasso procedure. While ZR-GLASSO and ZP-GLASSO are often not available in real applications, we use ZR-GLASSO and ZP-GLASSO as benchmarks for quantifying the information loss of the remaining estimators which are constructed based on the observed binary or mixed variables $\mathbf{X}$. We find that the CLIME estimator in Section 4.2 has the similar performance to the L-GLASSO estimator. Hence, we only present the results for L-GLASSO.

Define the number of false positives $\mathrm{FP}(\lambda)$ and true positives $\mathrm{TP}(\lambda)$ with regularization parameter $\lambda$ as the number of false positives and true positives in identifying vanishing entries in the lower off-diagonal elements. They are the number of lower off-diagonal elements $(i, j)$ such that $\Omega_{ij} = 0$ but the estimated $\Omega_{ij}$ is nonzero, and the number of lower off-diagonal elements $(i, j)$ such that $\Omega_{ij} \neq 0$ and the estimated $\Omega_{ij}$ is nonzero. Define the false positive rate $\mathrm{FPR}(\lambda)$ and true positive rate $\mathrm{TPR}(\lambda)$ as

$$\mathrm{FPR}(\lambda) = \frac{\mathrm{FP}(\lambda)}{d(d-1)/2 - |E|} \quad \text{and} \quad \mathrm{TPR}(\lambda) = \frac{\mathrm{TP}(\lambda)}{|E|},$$

where $|E|$ is the number of non-vanishing lower off-diagonal elements. Figure 1 shows the plot of $\mathrm{FPR}(\lambda)$ against $\mathrm{TPR}(\lambda)$ for L-GLASSO, L-GSCAD, AMLE, ZR-GLASSO and ZP-GLASSO, when $n = 200$. The naive method has similar ROC curves to AMLE and is omitted in the figure for clarification purposes. We find that L-GLASSO always yields higher TPR than AMLE for any fixed

FPR under all four scenarios, and L-GSCAD improves L-GLASSO in terms of graph recovery. By comparing the ROC curves in Scenarios (a) and (b), L-GLASSO and L-GSCAD are more robust to the data misclassification than the benchmark estimators ZR-GLASSO and ZP-GLASSO. This robustness property demonstrates the advantage of the dichotomization method. In the absence of misclassification, it is seen that the ROC curves of L-GLASSO and ZR-GLASSO are similar, suggesting little information loss for graph recovery due to the dichotomization procedure.

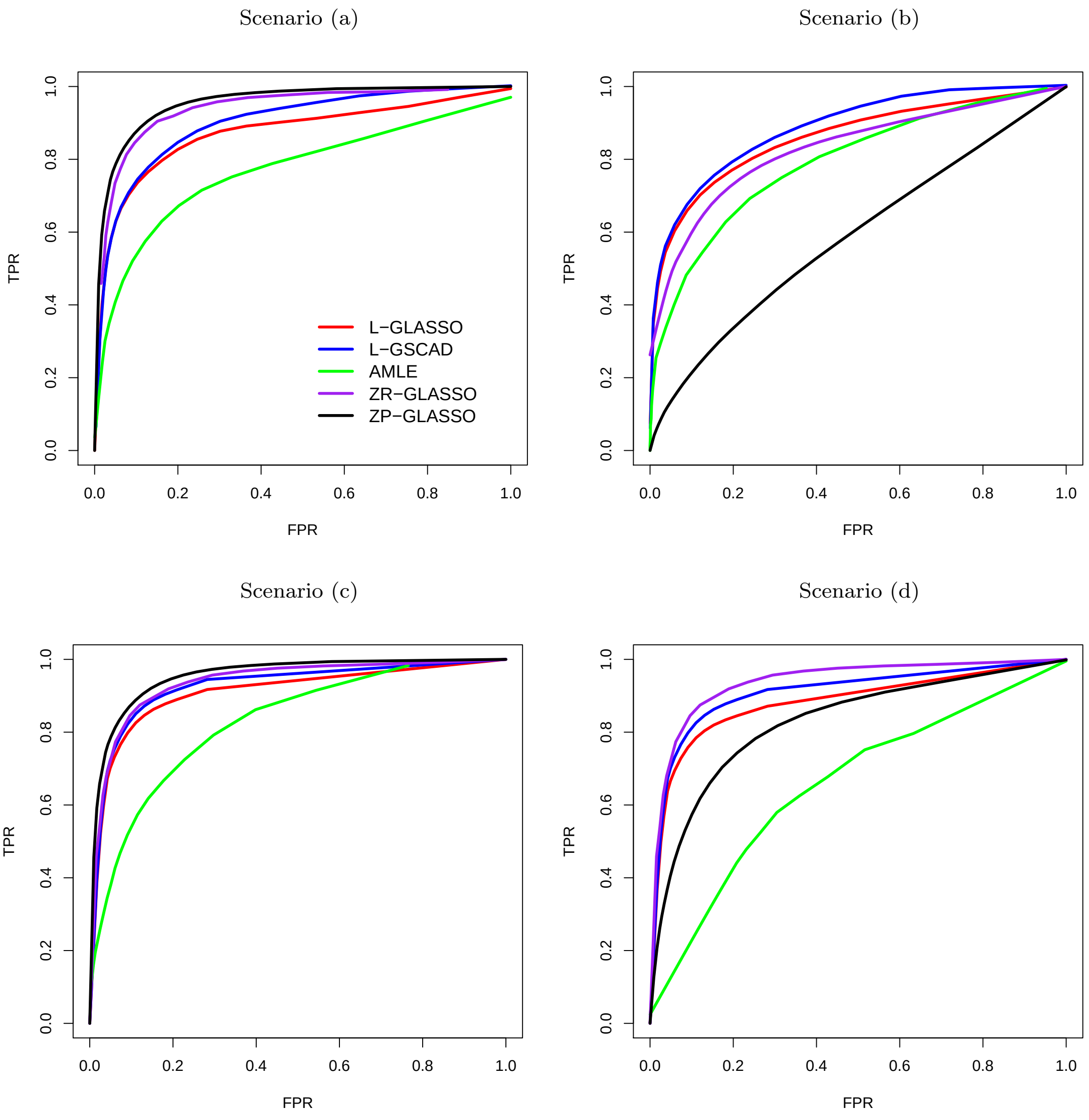

Figure 1: Plot of FPR against TPR for graph recovery under the four scenarios, when $n = 200$.

Table 1 reports the mean estimation error of $\widehat{\mathbf{\Omega}} - \mathbf{\Omega}$ in terms of the Frobenius and the spectral norms. The entries under the Frobenius and the spectral norms are calculated at the oracle regularization parameters $\lambda_F^*$ and $\lambda_s^*$ given by $\lambda_F^* = \text{argmin}_\lambda ||\widehat{\mathbf{\Omega}} - \mathbf{\Omega}||_F$, and $\lambda_s^* = \text{argmin}_\lambda ||\widehat{\mathbf{\Omega}} - \mathbf{\Omega}||_s$, respectively. It is seen that L-GLASSO has smaller estimation error than AMLE and the naive method under all scenarios. This phenomenon becomes more transparent, as $n$ increases. In ad-dition, L-GSCAD further reduces the estimation error of L-GLASSO. By comparing Scenarios (a) and (b), we find that the misclassification has little impact on L-GLASSO and L-GSCAD but significantly inflates the estimation error of ZR-GLASSO and ZP-GLASSO. The results under Scenarios (c) and (d) suggest that L-GLASSO and L-GSCAD are also robust to the non-Gaussian assumption on the continuous components. In contrast, AMLE and the native method have much larger estimation error under the non-Gaussian assumption in Scenario (d) than that in Scenario (c). Under the scenarios without misclassification, L-GLASSO produces slightly larger estimation error than the benchmark estimator ZR-GLASSO.

Table 1: The average estimation error of L-GLASSO, L-GSCAD, AMLE, the naive method, ZR-GLASSO and ZP-GLASSO for $\widehat{\mathbf{\Omega}} - \mathbf{\Omega}$ as measured by the spectral (S) and the Frobenius (F) norms. Numbers in parentheses are the simulation standard errors.

| $n$ | Scenario | Norm | L-GLASSO | L-GSCAD | AMLE | Naive | ZR-GLASSO | ZP-GLASSO |
|---|---|---|---|---|---|---|---|---|
| 50 | (a) | F | 5.01(1.05) | 4.64(0.92) | 5.46(1.40) | 5.42(1.36) | 4.55(0.81) | 4.20(0.71) |
| | | S | 1.97(0.51) | 1.70(0.35) | 2.65(0.95) | 2.63(0.92) | 1.66(0.38) | 1.59(0.33) |
| | (b) | F | 5.09(0.74) | 4.71(0.66) | 5.31(0.90) | 5.28(0.88) | 4.95(0.72) | 5.24(0.84) |
| | | S | 2.03(0.40) | 1.79(0.45) | 2.58(0.58) | 2.52(0.60) | 1.93(0.42) | 2.20(0.48) |
| | (c) | F | 4.71(0.94) | 4.43(0.78) | 5.06(1.33) | 5.02(1.20) | 4.55(0.81) | 4.20(0.71) |
| | | S | 1.95(0.58) | 1.63(0.40) | 2.46(0.90) | 2.34(0.87) | 1.66(0.38) | 1.59(0.33) |
| | (d) | F | 5.03(1.02) | 4.58(0.89) | 8.08(1.83) | 8.99(2.13) | 4.70(0.76) | 5.21(1.02) |
| | | S | 2.05(0.62) | 1.77(0.53) | 4.18(1.56) | 4.88(1.88) | 1.68(0.39) | 2.07(0.55) |
| 100 | (a) | F | 4.37(0.59) | 4.01(0.49) | 4.76(0.80) | 4.56(0.63) | 3.78(0.43) | 3.60(0.39) |
| | | S | 1.71(0.36) | 1.53(0.30) | 2.39(0.63) | 1.91(0.42) | 1.39(0.24) | 1.29(0.21) |
| | (b) | F | 4.75(0.87) | 4.19(0.67) | 5.03(1.18) | 4.91(0.99) | 4.69(0.88) | 5.29(1.14) |
| | | S | 1.95(0.53) | 1.70(0.47) | 2.47(0.80) | 2.08(0.66) | 1.95(0.56) | 2.35(0.74) |
| | (c) | F | 4.25(0.58) | 3.93(0.50) | 4.45(0.66) | 4.39(0.56) | 3.78(0.43) | 3.60(0.39) |
| | | S | 1.66(0.41) | 1.42(0.27) | 2.11(0.56) | 1.80(0.40) | 1.39(0.24) | 1.29(0.21) |
| | (d) | F | 4.33(0.61) | 3.95(0.48) | 5.19(0.80) | 5.18(0.79) | 3.82(0.46) | 4.78(0.63) |
| | | S | 1.78(0.44) | 1.43(0.24) | 2.30(0.60) | 2.24(0.55) | 1.40(0.26) | 1.90(0.42) |
| 200 | (a) | F | 3.63(0.62) | 3.35(0.59) | 4.79(1.02) | 4.24(0.75) | 3.02(0.42) | 2.87(0.38) |
| | | S | 1.50(0.42) | 1.22(0.30) | 2.41(0.71) | 1.83(0.50) | 1.08(0.20) | 1.06(0.25) |
| | (b) | F | 3.86(0.59) | 3.49(0.34) | 4.80(0.86) | 4.48(0.69) | 3.78(0.43) | 5.06(0.82) |
| | | S | 1.64(0.42) | 1.25(0.22) | 2.38(0.61) | 1.96(0.49) | 1.39(0.24) | 2.37(0.58) |
| | (c) | F | 3.40(0.67) | 3.11(0.60) | 4.26(0.87) | 3.87(0.64) | 3.02(0.42) | 2.87(0.38) |
| | | S | 1.47(0.47) | 1.28(0.39) | 2.13(0.49) | 1.61(0.43) | 1.08(0.20) | 1.06(0.25) |
| | (d) | F | 3.49(0.73) | 3.14(0.55) | 4.91(0.83) | 4.82(0.80) | 3.05(0.42) | 4.37(0.64) |
| | | S | 1.60(0.55) | 1.25(0.34) | 2.19(0.60) | 2.05(0.50) | 1.08(0.22) | 1.93(0.51) |

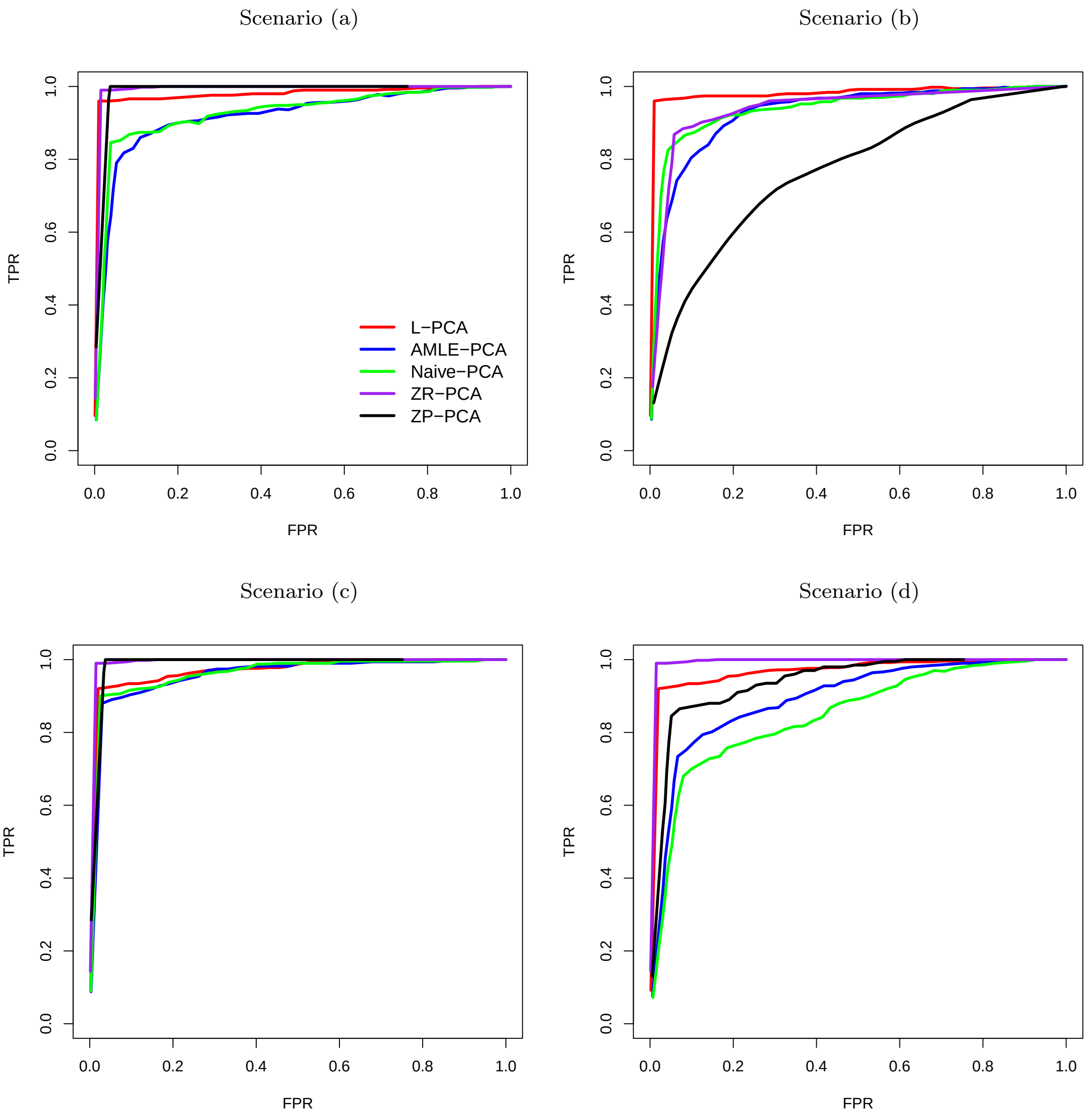

Figure 2: Plot of FPR against TPR for the recovery of the sparse leading eigenvector under the four scenarios, when $n = 200$.

## 7.2 Simulation Results for Principal Component Analysis

To evaluate the accuracy of eigenvector estimation, we use a different procedure to generate the correlation matrix $\mathbf{\Sigma}$. Consider two $d$ dimensional vectors $\boldsymbol{u}_1$ and $\boldsymbol{u}_2$, where

$$u_{1j} = \begin{cases} \frac{1}{\sqrt{10}} & 1 \leq j \leq 10 \\ 0 & \text{otherwise} \end{cases} \quad \text{and} \quad u_{2j} = \begin{cases} \frac{1}{\sqrt{10}} & 11 \leq j \leq 20 \\ 0 & \text{otherwise} \end{cases} .$$

Set

$$\widetilde{\mathbf{\Sigma}} := \sum_{j=1}^{2} (\omega_j - 1)\boldsymbol{u}_j\boldsymbol{u}_j^T + \mathbf{I}_d, \quad \text{where } \omega_1 = 5, \omega_2 = 4.$$

Table 2: The average estimation error of L-PCA, AMLE-PCA, Naive-PCA, ZR-PCA and ZP-PCA as measured by $\sin^2 \angle(\widehat{\boldsymbol{v}}_1, \boldsymbol{v}_1)$. Numbers in parentheses are the simulation standard errors.

| $n$ | Scenario | L-PCA | AMLE-PCA | Naive-PCA | ZR-PCA | ZP-PCA |
|---|---|---|---|---|---|---|
| 50 | (a) | 0.56(0.16) | 0.82(0.19) | 0.78(0.21) | 0.36(0.22) | 0.34(0.20) |
| | (b) | 0.60(0.15) | 0.85(0.15) | 0.83(0.16) | 0.66(0.17) | 0.92(0.12) |
| | (c) | 0.41(0.21) | 0.49(0.32) | 0.57(0.28) | 0.36(0.22) | 0.34(0.20) |
| | (d) | 0.50(0.23) | 0.72(0.22) | 0.79(0.20) | 0.39(0.19) | 0.77(0.14) |
| 100 | (a) | 0.32(0.22) | 0.66(0.26) | 0.59(0.29) | 0.15(0.17) | 0.13(0.10) |
| | (b) | 0.40(0.20) | 0.76(0.20) | 0.69(0.26) | 0.43(0.22) | 0.89(0.11) |
| | (c) | 0.21(0.20) | 0.32(0.38) | 0.32(0.38) | 0.15(0.17) | 0.13(0.10) |
| | (d) | 0.30(0.27) | 0.68(0.29) | 0.74(0.27) | 0.19(0.21) | 0.62(0.27) |
| 200 | (a) | 0.18(0.23) | 0.38(0.28) | 0.27(0.30) | 0.09(0.10) | 0.08(0.05) |
| | (b) | 0.23(0.27) | 0.55(0.31) | 0.41(0.33) | 0.18(0.16) | 0.84(0.15) |
| | (c) | 0.12(0.21) | 0.19(0.30) | 0.17(0.28) | 0.09(0.10) | 0.08(0.05) |
| | (d) | 0.15(0.23) | 0.37(0.34) | 0.37(0.32) | 0.11(0.15) | 0.53(0.39) |

The latent correlation matrix $\boldsymbol{\Sigma}$ is $\boldsymbol{\Sigma} = \text{diag}(\widetilde{\boldsymbol{\Sigma}})^{-1/2} \cdot \widetilde{\boldsymbol{\Sigma}} \cdot \text{diag}(\widetilde{\boldsymbol{\Sigma}})^{-1/2}$. Same as Section 7.1, we consider the simulation scenarios (a)–(d).

For each simulated data set, we compare the following five methods to calculate the leading eigenvector: (1) L-PCA, the truncated power method based on $\widehat{\mathbf{R}}$; (2) AMLE-PCA, the same method based on the modified sample covariance matrix in Banerjee et al. (2008); (3) Naive-PCA, the same method based on the sample covariance estimator; (4) ZR-PCA, the truncated power method with the rank based correlation estimator of $\text{cov}(\boldsymbol{Z})$ as the input and (5) ZP-PCA, the truncated power method with the sample correlation estimator of $\text{cov}(\boldsymbol{Z})$ as the input. In ZR-PCA and ZP-PCA, we assume that $\boldsymbol{Z}$ is observed. Again, ZR-PCA and ZP-PCA serve as benchmarks for evaluating the information loss of the PCA methods (L-PCA, AMLE-PCA and Naive-PCA) for dichotomized data.

For the estimation of leading eigenvector, define the number of false positives $\text{FP}(k)$ and true positives $\text{TP}(k)$ with regularization parameter $k$ as the number of components $i$ such that $v_i = 0$ and the estimated $v_i$ is nonzero, and the number of components $i$ such that $v_i \neq 0$ and the estimated $v_i$ is nonzero. Define the false positive rate $\text{FPR}(k)$ and true positive rate $\text{TPR}(k)$ to be

$$\text{FPR}(k) = \frac{\text{FP}(k)}{d - |E|} \quad \text{and} \quad \text{TPR}(k) = \frac{\text{TP}(\lambda)}{|E|},$$

where $|E|$ is the number of components $i$ such that $v_i \neq 0$. Figure 2 shows the plot of $\text{FPR}(k)$ against $\text{TPR}(k)$ for L-PCA, AMLE-PCA, Naive-PCA, ZR-PCA and ZP-PCA, when $n = 200$. It is easily seen that L-PCA outperforms AMLE-PCA and Naive-PCA in terms of feature selection under all four scenarios, and is only slightly inferior to the benchmark estimator ZR-PCA.

To evaluate the estimation accuracy of the leading eigenvector, we compare L-PCA, AMLE-PCA, Naive-PCA, ZR-PCA and ZP-PCA at $k = 10$. Table 2 shows the averaged distance between the estimated leading eigenvector $\widehat{\boldsymbol{v}}_1$ with the truth for all four estimation methods. Similar to the graph estimation, we find that our method (L-PCA) has smaller estimation error than AMLE-PCA and Naive-PCA under all scenarios, and is robust to the misclassification of the binary data and the non-Gaussian assumption for the continuous data. In Scenarios (a), (c) and (d), compared with ZR-PCA, the information loss of L-PCA is moderate. In Scenario (b), L-PCA even shows some advantage over ZR-PCA and ZP-PCA. This is because L-PCA is less affected than ZR-PCA and ZP-PCA by the extreme values in $\boldsymbol{Z}$.

### 7.3 Analysis of Arabidopsis Data

In this section, we consider the graph estimation and PCA for the Arabidopsis data set analyzed by Lange and Ghassemian (2003); Wille et al. (2004); Ma et al. (2007). As an illustration, we focus on 39 genes which are possibly related to the mevalonate or non-mevalonate pathway. In addition, 118 GeneChip (Affymetrix) microarrays are used to measure the gene expression values under various experimental conditions.

In the analysis of gene expression data collected from different platforms, there often exists unwanted variation among different experiments known as the batch effects in the literature (McCall et al., 2014; Lazar et al., 2013). To remove the batch effects in the Arabidopsis data, we apply the adaptive dichotomization method implemented by the `ArrayBin` package in `R`. This method transforms the numerical expression data into 0/1 binary data, where genes with high expression values are encoded as 1 and the genes with lower expression values are encoded as 0. Although the information loss is inevitable in the discretization procedure, McCall and Irizarry (2011) argued that this procedure can potentially improve the accuracy of the statistical analysis. In contrast to Wille et al. (2004); Ma et al. (2007) which imposed the Gaussian model assumption to the numerical expression values, we work on the derived binary data with the purpose of removing the batch effects.

We first compare the performance of graph recovery based on several methods. To ensure a fair comparison, we only consider the estimators with the Lasso penalty, i.e., L-GLASSO, AMLE and the naive method described in Section 7.1. The turning parameters are selected separately, such that the number of edges in the estimated graphs by all three methods are identical. The number of different edges among L-GLASSO and AMLE, and among L-GLASSO and the naive method is presented in Table 3. The estimated graphs by the AMLE and the naive method are very similar, which is consistent with our findings in the simulation studies. More importantly, we find that our estimator produces $30\% \sim 60\%$ different edges compared to the AMLE and the naive method, depending on the sparsity level of the estimated graphs. From a biological perspective, many important association patterns are identified by L-GLASSO rather than AMLE and the naive method. For instance, when the number of total edges is 10, L-GLASSO is the only method that concludes that genes CMK and MCT, and CMK and MECPS are dependent. Both of these genes are on the non-mevalonate pathway and are known to be associated in the literature (Hsieh and Goodman, 2005; Phillips et al., 2008; Ruiz-Sola and Rodríguez-Concepción, 2012). Similarly, the association between genes MECPS and HDS supported by Phillips et al. (2008) is recovered by L-GLASSO rather than AMLE and the naive method. Hence, the gene dependence graph estimated by our method seems to be more biologically meaningful than that estimated by AMLE and the naive method.

To perform dimension reduction, we apply the L-PCA, AMLE-PCA and Naive-PCA described in Section 7.2 to the data set. The genes in the data set belong to either the mevalonate group or the non-mevalonate group. Given a sparsely estimated eigenvector, we can calculate the matches with the mevalonate group as the number of genes in this group that are selected by the sparse eigenvector. The matches with the non-mevalonate group can be similarly defined. We compare the performance of the three PCA methods based on the match rate for the estimated leading eigenvector, which is defined as the ratio of matches to the cardinality of the estimated support sets. Since the sum of the match rate with these two groups is 1, larger values are kept. Table 4 reports the match rate of L-PCA, AMLE-PCA and Naive-PCA for different cardinality of the estimated support sets. We find that under all considered scenarios for $k$, the match rate of L-PCA remains above 80%, which is significantly larger than that of AMLE-PCA and Naive-PCA. This implies that the genes identified by L-PCA tend to belong to either the mevalonate group or the non-mevalonate group. In contrast, AMLE-PCA and Naive-PCA may fail to follow this group structure. Therefore, as a dimension reduction procedure, L-PCA finds a small number of genes characterizing the major variation pattern in the data, and meanwhile these genes are more likely to interact in the same biological pathway.

Table 3: The number of different edges among L-GLASSO and AMLE, and among L-GLASSO and the naive method in the Arabidopsis data.

| | L-GLASSO vs AMLE | | | | | L-GLASSO vs Naive | | | | |
|---|---|---|---|---|---|---|---|---|---|---|
| Number of total edges | 80 | 60 | 45 | 25 | 10 | 80 | 60 | 45 | 25 | 10 |
| Number of different edges | 27 | 19 | 15 | 7 | 6 | 27 | 18 | 14 | 7 | 6 |

Table 4: The match rates of L-PCA, AMLE-PCA and Naive-PCA for the estimated leading eigenvectors under different cardinality of the estimated support sets $k$, in the Arabidopsis data.

| k | L-PCA | AMLE-PCA | Naive-PCA |
|---|---|---|---|
| 5 | 1.00 | 0.80 | 0.80 |
| 7 | 0.86 | 0.50 | 0.50 |
| 10 | 0.80 | 0.50 | 0.50 |
| 12 | 0.83 | 0.50 | 0.50 |
| 15 | 0.80 | 0.54 | 0.60 |

# 8 Discussion

In this paper, we propose a latent Gaussian copula model for binary data and mixed data. Our model is fundamentally different from the existing ones in the literature in the sense that the dependence structure and variation patterns in the unobserved latent variables are of primary interest. A convenient rank based approach is exploited to estimate the latent graph and conduct PCA for the latent random variables. The theoretical properties for both graph estimation and PCA are established.

In our PCA procedure, we confine our attention to the estimation of the leading eigenvector. If the subspace spanned by the top $m$ leading eigenvectors of $\Sigma$ is of interest, we can use the Fantope projection approach (Vu et al., 2013), which is the extension of the convex relaxation method in Section 5.1, to estimate the sparse principal subspace. A similar truncated power method can be used to further refine the estimator.

Although we focus on the binary data in this paper, our methods can be extended to the discrete data with more than two categories. For instance, once the Kendall's tau is defined, we can follow the similar arguments in (3) to derive the bridge function that connects the latent correlation matrix to the population version of Kendall's tau. The properties of this procedure are left for future investigation.

# Acknowledgement

Fan's research was supported by NIH (National Institutes of Health) Grant 2R01-GM072611-9 and NSF (National Science Foundation) Grant DMS-1206464. Liu's research was supported by NSF Grants III-1116730 and III-1332109, NIH Grants R01MH102339, R01GM083084, and R01HG06841, and FDA Grant HHSF223201000072C. Zou's research was supported in part by NSF grant DMS-0846068.

# A Appendix

## A.1 Proof of Lemma 3.1

*Proof.* We will show that the partial derivative of $F(t;\Delta_j,\Delta_k)$ with respect to $t$ is positive, i.e., $\partial F(t;\Delta_j,\Delta_k)/\partial t > 0$.

To show this result, we first note that, for a bivariate random variable $(X_j, X_k)$ with distribution function $\Phi_2(\cdot,\cdot,t)$, the conditional distribution satisfies

$$X_k|X_j = x_j \sim N(tx_j, (1-t^2)).$$

Then,

$$\Phi_2(\Delta_j,\Delta_k,t) = \int_{-\infty}^{\Delta_j} \Phi\left(\frac{\Delta_k - tx}{\sqrt{1-t^2}}\right)\phi(x)dx, \tag{24}$$

where $\phi(x)$ is the probability density function of a standard normal variable. Hence,

$$\frac{\partial F(t;\Delta_j,\Delta_k)}{\partial t} = 2\frac{\partial}{\partial t}\int_{-\infty}^{\Delta_j} \Phi\left(\frac{\Delta_k - tx}{\sqrt{1-t^2}}\right)\phi(x)dx. \tag{25}$$

Since $\Phi(x) < 1$, from the dominated convergence theorem, it is valid to interchange the differentiation and integration in equation (25). We obtain,

$$\frac{\partial F(t;\Delta_j,\Delta_k)}{\partial t} = 2\int_{-\infty}^{\Delta_j} \phi\left(\frac{\Delta_k - tx}{\sqrt{1-t^2}}\right)\phi(x)\frac{-x+t\Delta_k}{(1-t^2)^{3/2}}dx. \tag{26}$$

If $\Delta_j < t\Delta_k$, the integrand in the right hand side of (26) is positive, and hence $\partial F(t)/\partial t > 0$. If $\Delta_j \geq t\Delta_k$, the integral in the right hand side of equation (26) is a decreasing function of $\Delta_j$. This entails that

$$
\begin{aligned}
\frac{\partial F(t;\Delta_j,\Delta_k)}{\partial t} &> 2\int_{-\infty}^{\infty} \phi\left(\frac{\Delta_k - tx}{\sqrt{1-t^2}}\right)\phi(x)\frac{-x+t\Delta_k}{(1-t^2)^{3/2}}dx \\
&= 2\frac{\partial}{\partial t}\int_{-\infty}^{\infty} \Phi\left(\frac{\Delta_k - tx}{\sqrt{1-t^2}}\right)\phi(x)dx \\
&= 2\frac{\partial}{\partial t}\Phi_2(\infty,\Delta_k,t) \\
&= 2\frac{\partial}{\partial t}\Phi(\Delta_k) = 0,
\end{aligned} \tag{27}
$$

where the first equality follows from interchanging differentiation and integration, and the second equality follows from equation (24). Hence $F(t,\Delta_j,\Delta_k)$ is strictly increasing with $t$. □

## A.2 Proof of Lemma 3.2

*Proof.* It suffices to show that there exists a constant $L_2$ such that $\partial F^{-1}(\tau;\Delta_j,\Delta_k)/\partial\tau < L_2$, which is equivalent to $\partial F(t;\Delta_j,\Delta_k)/\partial t > 1/L_2$ for all $|t| \leq 1-\delta$. We separate this into two cases.

For the case that $\Delta_j < t\Delta_k$, the integrand is nonnegative in (26) and hence

$$
\frac{\partial F(t;\Delta_j,\Delta_k)}{\partial t} \geq 2\int_{-\infty}^{\Delta_j} \phi\left(\frac{\Delta_k - tx}{\sqrt{1-t^2}}\right)\phi(x)(-x+t\Delta_k)dx.
$$

With $\eta = \min\{-|t\Delta_k| - 1, \Delta_j\}$, when $x < \eta$, we have $-x + t\Delta_k \geq 1$. Therefore, the derivative is further bounded from below by

$$
\begin{aligned}
2\int_{-\infty}^{\eta} \phi\left(\frac{\Delta_k - tx}{\sqrt{1-t^2}}\right)\phi(x)dx &\geq 2\int_{-\infty}^{\eta} \phi\left(\frac{M+|x|}{\sqrt{2\delta-\delta^2}}\right)\phi(x)dx \\
&\geq 2\int_{-\infty}^{-M-1} \phi\left(\frac{M+|x|}{\sqrt{2\delta-\delta^2}}\right)\phi(x)dx \equiv \frac{1}{L'}.
\end{aligned}
$$

where the first inequality follows from

$$
\frac{|\Delta_k - tx|}{\sqrt{1-t^2}} \leq \frac{|\Delta_k| + |t||x|}{\sqrt{1-t^2}} \leq \frac{M+|x|}{\sqrt{2\delta-\delta^2}}
$$

for all $|t| \leq 1-\delta$ and second inequality follows from the fact that $\eta > -M-1$.

We now consider the case that $\Delta_j \geq t\Delta_k$. By equation (27), we have

$$
\int_{-\infty}^{\Delta_j} \phi\left(\frac{\Delta_k - tx}{\sqrt{1-t^2}}\right)\phi(x)\frac{-x+t\Delta_k}{(1-t^2)^{3/2}}dx = -\int_{\Delta_j}^{\infty} \phi\left(\frac{\Delta_k - tx}{\sqrt{1-t^2}}\right)\phi(x)\frac{-x+t\Delta_k}{(1-t^2)^{3/2}}dx.
$$

With the change of the variable $u = x - t\Delta_k$, the above integral can be written as

$$
\int_{\Delta_j - t\Delta_k}^{\infty} \phi\left(\frac{(1-t^2)\Delta_k - tu}{\sqrt{1-t^2}}\right)\phi(u+t\Delta_k)\frac{u}{(1-t^2)^{3/2}}du,
$$

This and the fact that

$$\frac{|(1-t^2)\Delta_k - tx|}{\sqrt{1-t^2}} \le M + \frac{|x|}{\sqrt{2\delta-\delta^2}}.$$

entail

$$\begin{aligned}\frac{\partial F(t;\Delta_j,\Delta_k)}{\partial t} &\ge 2\int_{2M}^{\infty}\phi\left(\frac{(1-t^2)\Delta_k - tx}{\sqrt{1-t^2}}\right)\phi(x+t\Delta_k)\frac{x}{(1-t^2)^{3/2}}dx\\ &\ge 2\int_{2M}^{\infty}\phi\left(M+\frac{|x|}{\sqrt{2\delta-\delta^2}}\right)\phi(x+M)xdx \equiv \frac{1}{L''}.\end{aligned}$$

Then we can take $L_2 = \max\{L', L''\}$, and $L_2$ is independent of $\Delta_j$ and $\Delta_k$. This completes the proof. □

## A.3 Proof of Theorem 3.1

**Lemma A.1.** *$\Phi^{-1}(y)$ is Lipschitz in $y \in [\Phi(-2M), \Phi(2M)]$, i.e., there exists a Lipschitz constant $L_1$ such that*

$$|\Phi^{-1}(y_1) - \Phi^{-1}(y_2)| \le L_1|y_1 - y_2|.$$

*Proof of Lemma A.1.* It suffices to show that there exists a constant $L_2$ such that $d\Phi^{-1}(y)/dy < L_1$, which is equivalent to $\phi(x) = d\Phi(x)/dx > 1/L_1$ for all $x \in [-2M, 2M]$. Apparently, this is true by taking $L_1 = 1/\phi(2M)$. □

*Proof of Theorem 3.1.* Note that $\widehat{\Delta}_j = \Phi^{-1}\left(1 - \frac{1}{n}\sum_{i=1}^n X_{ij}\right)$. By Lemma A.1, under the event $A_j = \{|\widehat{\Delta}_j| \le 2M\}$, we obtain

$$\begin{aligned}|\widehat{\Delta}_j - \Delta_j| &= \left|\Phi^{-1}\left(1-\frac{1}{n}\sum_{i=1}^n X_{ij}\right) - \Phi^{-1}(\Phi(\Delta_j))\right|\\ &\le L_1\left|\frac{1}{n}\sum_{i=1}^n X_{ij} - (1-\Phi(\Delta_j))\right|.\end{aligned}\tag{28}$$

The exception probability is controlled by

$$\begin{aligned}\mathbb{P}(A_j^c) &\le \mathbb{P}(|\widehat{\Delta}_j - \Delta_j| > M)\\ &\le \mathbb{P}\left(\left|\frac{1}{n}\sum_{i=1}^n X_{ij} - (1-\Phi(\Delta_j))\right| > \frac{M}{L_1}\right)\\ &\le 2\exp\left(-\frac{M^2 n}{2L_1^2}\right),\end{aligned}\tag{29}$$

where the last step follows from the Hoeffding's inequality, since $X_{ij}$ is binary. For any $t > 0$, the $(j,k)$th element of $\widehat{\mathbf{R}}$ satisfies

$$\begin{aligned}\mathbb{P}\left(|F^{-1}(\widehat{\tau}_{jk};\widehat{\Delta}_j,\widehat{\Delta}_k) - \Sigma_{jk}| > t\right) &\le \mathbb{P}\left(\{|F^{-1}(\widehat{\tau}_{jk};\widehat{\Delta}_j,\widehat{\Delta}_k) - \Sigma_{jk}| > t\} \cap A_j \cap A_k\right)\\ &\quad + \mathbb{P}(A_j^c) + \mathbb{P}(A_k^c).\end{aligned}$$

Note that $\Sigma_{jk} = F^{-1}(F(\Sigma_{jk}; \widehat{\Delta}_j, \widehat{\Delta}_k); \widehat{\Delta}_j, \widehat{\Delta}_k)$. From Lemma 3.2,

$$
\begin{aligned}
& \mathbb{P}\left(\{|F^{-1}(\widehat{\tau}_{jk}; \widehat{\Delta}_j, \widehat{\Delta}_k) - \Sigma_{jk}| > t\} \cap A_j \cap A_k\right) \\
\leq\ & \mathbb{P}\left(\{L_2|\widehat{\tau}_{jk} - F(\Sigma_{jk}; \widehat{\Delta}_j, \widehat{\Delta}_k)| > t\} \cap A_j \cap A_k\right) \\
\leq\ & \mathbb{P}\left(L_2|\widehat{\tau}_{jk} - F(\Sigma_{jk}; \Delta_j, \Delta_k)| > t/2\right) \\
& \quad + \mathbb{P}\left(\{L_2|F(\Sigma_{jk}; \widehat{\Delta}_j, \widehat{\Delta}_k) - F(\Sigma_{jk}; \Delta_j, \Delta_k)| > t/2\} \cap A_j \cap A_k\right) \\
\equiv\ & I_1 + I_2. \qquad (30)
\end{aligned}
$$

Since $\widehat{\tau}_{jk}$ is a U-statistic with bounded kernel, the Hoeffding's inequality for U-statistics yields,

$$
I_1 \leq 2\exp\left(-\frac{nt^2}{8L_2^2}\right). \qquad (31)
$$

Let $\Phi_{21}(x, y, t) = \partial\Phi_2(x, y, t)/\partial x$, and $\Phi_{22}(x, y, t) = \partial\Phi_2(x, y, t)/\partial y$. For $I_2$, we have

$$
\begin{aligned}
& \left|F(\Sigma_{jk}; \widehat{\Delta}_j, \widehat{\Delta}_k) - F(\Sigma_{jk}; \Delta_j, \Delta_k)\right| \\
\leq\ & 2\left|\Phi_2(\widehat{\Delta}_j, \widehat{\Delta}_k, \Sigma_{jk}) - \Phi_2(\Delta_j, \Delta_k, \Sigma_{jk})\right| + 2\left|\Phi(\widehat{\Delta}_j)\Phi(\widehat{\Delta}_k) - \Phi(\Delta_j)\Phi(\Delta_k)\right| \\
\leq\ & 2\left|\Phi_{21}(\xi_1)(\widehat{\Delta}_j - \Delta_j)\right| + 2\left|\Phi_{22}(\xi_2)(\widehat{\Delta}_k - \Delta_k)\right| + 2\Phi(\widehat{\Delta}_k)\left|\phi(\xi_3)(\widehat{\Delta}_j - \Delta_j)\right| \\
& \quad + 2\Phi(\Delta_j)\left|\phi(\xi_4)(\widehat{\Delta}_k - \Delta_k)\right|, \qquad (32)
\end{aligned}
$$

where $\xi_1, \xi_2, \xi_3$ and $\xi_4$ are the intermediate values from the mean value theorem. It is easily seen that

$$
\Phi_{21}(x, y, t) = \frac{\partial}{\partial x}\int_{-\infty}^{x}\Phi\left(\frac{y - tz}{\sqrt{1-t^2}}\right)\phi(z)dz = \Phi\left(\frac{y - tx}{\sqrt{1-t^2}}\right)\phi(x) \leq \frac{1}{\sqrt{2\pi}}.
$$

Similarly, we can show that $\Phi_{22}(x, y, t) \leq \frac{1}{\sqrt{2\pi}}$. Then together with (32), we get

$$
\left|F(\Sigma_{jk}; \widehat{\Delta}_j, \widehat{\Delta}_k) - F(\Sigma_{jk}; \Delta_j, \Delta_k)\right| \leq 4\frac{1}{\sqrt{2\pi}}\left\{|\widehat{\Delta}_j - \Delta_j| + |\widehat{\Delta}_k - \Delta_k|\right\}. \qquad (33)
$$

Combining (30), (33) and (28), we find

$$
\begin{aligned}
I_2 \ &\leq\ \mathbb{P}\left(L_1\left|\frac{1}{n}\sum_{i=1}^{n}X_{ij} - (1 - \Phi(\Delta_j))\right| > \frac{t\sqrt{2\pi}}{16L_2}\right) + \mathbb{P}\left(L_1\left|\frac{1}{n}\sum_{i=1}^{n}X_{ik} - (1 - \Phi(\Delta_k))\right| > \frac{t\sqrt{2\pi}}{16L_2}\right) \\
&\leq\ 4\exp\left(-\frac{nt^2\pi}{16^2L_1^2L_2^2}\right), \qquad (34)
\end{aligned}
$$

where the last step follows from the Hoeffding's inequality. Combining results (31), (34) and (29), we now obtain

$$
\mathbb{P}\left(\left|F^{-1}(\widehat{\tau}_{jk}; \widehat{\Delta}_j, \widehat{\Delta}_k) - \Sigma_{jk}\right| > t\right) \leq 2\exp\left(-\frac{nt^2}{8L_2^2}\right) + 4\exp\left(-\frac{nt^2\pi}{16^2L_1^2L_2^2}\right) + 4\exp\left(-\frac{M^2n}{2L_1^2}\right).
$$

The bound on $\mathbb{P}\left(\sup_{j,k}\left|F^{-1}(\widehat{\tau}_{jk}; \widehat{\Delta}_j, \widehat{\Delta}_k) - \Sigma_{jk}\right| > t\right)$ follows from the union bound. Hence, taking $t = C\sqrt{\log d/n}$ for some constant $C$, $\sup_{j,k}|\widehat{R}_{jk} - \Sigma_{jk}| \leq C\sqrt{\log d/n}$ with probability greater than $1 - d^{-1}$. □

### A.4 Proof of Theorem 4.1

*Proof.* According to the definition of $\widehat{\mathbf{R}}_p$,

$$||\widehat{\mathbf{R}}_p - \mathbf{\Sigma}||_{\max} \le ||\widehat{\mathbf{R}}_p - \widehat{\mathbf{R}}||_{\max} + ||\widehat{\mathbf{R}} - \mathbf{\Sigma}||_{\max} \le 2||\widehat{\mathbf{R}} - \mathbf{\Sigma}||_{\max}.$$

We have

$$\mathbb{P}\left(||\widehat{\mathbf{R}}_p - \mathbf{\Sigma}||_{\max} \ge 2t\right) \le \mathbb{P}\left(||\widehat{\mathbf{R}} - \mathbf{\Sigma}||_{\max} \ge t\right).$$

This completes the proof by applying Theorem 3.1. □

### A.5 Proof of Theorem 4.2

*Proof.* The proof follows the same arguments as those in the proofs of Theorems 1, 4 and 7 in Cai et al. (2011). We omit the details. □

### A.6 Proof of Theorem 5.1

*Proof.* Let $\theta = \boldsymbol{v}_1^T \boldsymbol{v}$ for any given $\boldsymbol{v} \in \mathbb{S}^{d-1}$. Then, we can decompose

$$\boldsymbol{v} = \theta \boldsymbol{v}_1 + \sqrt{1-\theta^2}\boldsymbol{v}_1^*, \qquad \boldsymbol{v}_1^T \boldsymbol{v}_1^* = 0.$$

Using this,

$$\begin{aligned} \left\langle \mathbf{\Sigma}, \boldsymbol{v}_1\boldsymbol{v}_1^T - \boldsymbol{v}\boldsymbol{v}^T \right\rangle &= \lambda_1 - \boldsymbol{v}^T\mathbf{\Sigma}\boldsymbol{v} \\ &= \lambda_1 - [\theta^2\boldsymbol{v}_1^T\mathbf{\Sigma}\boldsymbol{v}_1 + (1-\theta^2)\boldsymbol{v}_1^{*T}\mathbf{\Sigma}\boldsymbol{v}_1^*] \\ &\le \lambda_1(1-\theta^2) - (1-\theta^2)\lambda_2. \end{aligned}$$

Together with the fact that $\sin^2\angle(\boldsymbol{v}_1, \boldsymbol{v}) = 1-\theta^2$, we have

$$\sin^2\angle(\boldsymbol{v}_1, \boldsymbol{v}) \le \frac{1}{\lambda_1 - \lambda_2}\left\langle \mathbf{\Sigma}, \boldsymbol{v}_1\boldsymbol{v}_1^T - \boldsymbol{v}\boldsymbol{v}^T \right\rangle. \tag{35}$$

By the definition of $\widehat{\boldsymbol{v}}_1(k)$, we have

$$\left\langle \widehat{\mathbf{R}}, \boldsymbol{v}_1\boldsymbol{v}_1^T - \widehat{\boldsymbol{v}}_1(k)\widehat{\boldsymbol{v}}_1(k)^T \right\rangle = \boldsymbol{v}_1^T\widehat{\mathbf{R}}\boldsymbol{v}_1 - \widehat{\boldsymbol{v}}_1(k)^T\widehat{\mathbf{R}}\widehat{\boldsymbol{v}}_1(k) \le 0. \tag{36}$$

Replacing $\boldsymbol{v}$ with $\widehat{\boldsymbol{v}}_1(k)$ in equation (35), by equation (36), we obtain

$$\begin{aligned} \sin^2\angle(\boldsymbol{v}_1, \widehat{\boldsymbol{v}}_1(k)) &\le \frac{1}{\lambda_1-\lambda_2}\left\langle \mathbf{\Sigma}, \boldsymbol{v}_1\boldsymbol{v}_1^T - \widehat{\boldsymbol{v}}_1(k)\widehat{\boldsymbol{v}}_1(k)^T \right\rangle \\ &\le \frac{1}{\lambda_1-\lambda_2}\left\langle \mathbf{\Sigma} - \widehat{\mathbf{R}}, \boldsymbol{v}_1\boldsymbol{v}_1^T - \widehat{\boldsymbol{v}}_1(k)\widehat{\boldsymbol{v}}_1(k)^T \right\rangle \\ &\le \frac{1}{\lambda_1-\lambda_2}||\operatorname{vec}(\widehat{\mathbf{R}} - \mathbf{\Sigma})||_\infty \cdot ||\operatorname{vec}(\boldsymbol{v}_1\boldsymbol{v}_1^T - \widehat{\boldsymbol{v}}_1(k)\widehat{\boldsymbol{v}}_1(k)^T)||_1. \end{aligned} \tag{37}$$

Note that $||\operatorname{vec}(\boldsymbol{v}_1\boldsymbol{v}_1^T - \widehat{\boldsymbol{v}}_1(k)\widehat{\boldsymbol{v}}_1(k)^T)||_0 \le k_0^2 + k^2 \le 2k^2$, and

$$\begin{aligned} ||\operatorname{vec}(\boldsymbol{v}_1\boldsymbol{v}_1^T - \widehat{\boldsymbol{v}}_1(k)\widehat{\boldsymbol{v}}_1(k)^T)||_2^2 &= \operatorname{Tr}\left\{(\boldsymbol{v}_1\boldsymbol{v}_1^T - \widehat{\boldsymbol{v}}_1(k)\widehat{\boldsymbol{v}}_1(k)^T)(\boldsymbol{v}_1\boldsymbol{v}_1^T - \widehat{\boldsymbol{v}}_1(k)\widehat{\boldsymbol{v}}_1(k)^T)\right\} \\ &= 2(1 - (\boldsymbol{v}_1^T\widehat{\boldsymbol{v}}_1(k))^2) = 2\sin^2\angle(\boldsymbol{v}_1, \widehat{\boldsymbol{v}}_1(k)). \end{aligned}$$

By (37) and the Cauchy-Schwartz inequality, we obtain

$$\sin^2 \angle(\boldsymbol{v}_1, \widehat{\boldsymbol{v}}_1(k)) \quad \le \quad \frac{1}{\lambda_1 - \lambda_2} || \operatorname{vec}(\widehat{\mathbf{R}} - \boldsymbol{\Sigma})||_\infty 2k |\sin \angle(\boldsymbol{v}_1, \widehat{\boldsymbol{v}}_1(k))|,$$

which implies that, with probability greater than $1 - d^{-1}$,

$$|\sin \angle(\boldsymbol{v}_1, \widehat{\boldsymbol{v}}_1(k))| \le \frac{2k}{\lambda_1 - \lambda_2} || \operatorname{vec}(\widehat{\mathbf{R}} - \boldsymbol{\Sigma})||_\infty \le \frac{2k}{\lambda_1 - \lambda_2} C \sqrt{\frac{\log d}{n}},$$

where the last step follows from Theorem 3.1. This completes the proof of (19).

The proof of (20) follows directly from Theorem 1 in Yuan and Zhang (2013) and the proof is omitted. □

## A.7 Proof of Theorem 5.2

*Proof.* Without loss of generality, we assume that $\boldsymbol{v}_1^T \widehat{\boldsymbol{v}}_1(k) \ge 0$. Then,

$$\sin^2 \angle(\widehat{\boldsymbol{v}}_1(k), \boldsymbol{v}_1) = 1 - (\boldsymbol{v}_1^T \widehat{\boldsymbol{v}}_1(k))^2 \ge 1 - \boldsymbol{v}_1^T \widehat{\boldsymbol{v}}_1(k) = \frac{||\boldsymbol{v}_1 - \widehat{\boldsymbol{v}}_1(k)||_2^2}{2}. \tag{38}$$

If $V_1 \not\subset \widehat{V}_1(k)$, there exists $j \in V_1 \cap \widehat{V}_1(k)^c$. Then, we have

$$||\boldsymbol{v}_1 - \widehat{\boldsymbol{v}}_1(k)||_2 \ge |v_{1j} - \widehat{v}_{1j}(k)| = |v_{1j}| \ge \min_{j \in V_1} |v_{1j}| \ge \frac{2\sqrt{2}Ck}{\lambda_1 - \lambda_2} \sqrt{\frac{\log d}{n}}. \tag{39}$$

Combing (38) and (39), we obtain

$$\sin \angle(\widehat{\boldsymbol{v}}_1(k), \boldsymbol{v}_1) \ge \frac{2Ck}{\lambda_1 - \lambda_2} \sqrt{\frac{\log d}{n}}.$$

By Theorem 5.1, the probability of $V_1 \not\subset \widehat{V}_1(k)$ is less than $1/d$.

Following the same arguments, we can show that $\mathbb{P}(V_1 \subset \widetilde{V}_1(k)) \ge 1 - d^{-1}$. □

## A.8 Proof of Lemma 6.3

*Proof.* Let $U_{ij} = f_j(Z_{ij})$ and $V_{ik} = f_k(X_{ik})$. Note that $(U_{ij}, (V_{ik} - V_{i'k}))/\sqrt{2})$ is standard bivariate normally distributed with correlation $\Sigma_{jk}/\sqrt{2}$, and $(U_{i'j}, (V_{ik} - V_{i'k}))/\sqrt{2})$ is standard bivariate normally distributed with correlation $-\Sigma_{jk}/\sqrt{2}$. By definition, $\tau_{jk}$ is given by

$$\begin{aligned} & E\{I(Z_{ij} > C_j)\operatorname{sign}(X_{ik} - X_{i'k})\} - E\left\{I(Z_{i'j} > C_j)\operatorname{sign}(X_{ik} - X_{i'k})\right\} \\ = \; & E\left[I(U_{ij} > \Delta_j)\operatorname{sign}(V_{ik} - V_{i'k})\right] - E\left[I(U_{i'j} > \Delta_j)\operatorname{sign}(V_{ik} - V_{i'k})\right] \end{aligned}$$

Using $\operatorname{sign}(x) = 2I(x > 0) - 1$, it follows from the definition of $\Phi_2$ that the above expectation can further be expressed as

$$\begin{aligned} & 2E\{I(U_{ij} > \Delta_j, V_{ik} - V_{i'k} > 0)\} - 2E\left\{I(U_{i'j} > \Delta_j, V_{ik} - V_{i'k} > 0)\right\} \\ = \; & 2\Phi_2(\Delta_j, 0, \Sigma_{jk}/\sqrt{2}) - 2\Phi_2(\Delta_j, 0, -\Sigma_{jk}/\sqrt{2}) \\ = \; & 4\Phi_2(\Delta_j, 0, \Sigma_{jk}/\sqrt{2}) - 2\Phi(\Delta_j), \end{aligned}$$

where the last step follows from $\Phi_2(\Delta_j, 0, -t) = \Phi(\Delta_j) - \Phi_2(\Delta_j, 0, t)$. When $\Delta_j = 0$, by the Sheppard's theorem (Sheppard, 1899), we get

$$H(t; 0) = 4\Phi_2(0, 0, t/\sqrt{2}) - 1 = \frac{2}{\pi}\sin^{-1}(t/\sqrt{2}).$$

By the proof of Lemma 3.1, we can show that

$$\frac{\partial H(t; \Delta_j)}{\partial t} = 4\frac{\partial \Phi_2(\Delta_j, 0, t/\sqrt{2})}{\partial t} > 0.$$

This implies that $H(t; \Delta_j)$ is strictly increasing with $t$, and therefore invertible. □

## A.9 Proof of Theorem 6.1

**Lemma A.2.** *Under Conditions (A1') and (A2'), $H^{-1}(\tau; \Delta_j)$ is Lipschitz in $\tau$ uniformly over $|\Delta_j| \le M$ and $|\tau| \le 1 - \delta$. Namely, there exists a Lipschitz constant $L_3$ such that*

$$|H^{-1}(\tau_1; \Delta_j) - H^{-1}(\tau_2; \Delta_j)| \le L_3|\tau_1 - \tau_2|,$$

*for all $|\Delta_j| \le M$ and $|\tau| \le 1 - \delta$.*

*Proof of Lemma A.2.* The proof follows the same argument as that for Lemma 3.2. We omit the details. □

*Proof of Theorem 6.1.* Let $A_j = \{|\widehat{\Delta}_j| \le 2M\}$. As shown in the proof of Theorem 3.1, by (29), we have

$$\mathbb{P}(A_j^c) \le 2\exp\left(-\frac{M^2 n}{2L_1^2}\right). \tag{40}$$

For any $t > 0$, we have

$$\mathbb{P}\left(|H^{-1}(\widehat{\tau}_{jk}; \widehat{\Delta}_j) - \Sigma_{jk}| > t\right) \le \mathbb{P}\left(\{|H^{-1}(\widehat{\tau}_{jk}; \widehat{\Delta}_j) - \Sigma_{jk}| > t\} \cap A_j\right) + \mathbb{P}(A_j^c). \tag{41}$$

By Lemma A.2,

$$\begin{aligned}
& \mathbb{P}\left(\{|H^{-1}(\widehat{\tau}_{jk}; \widehat{\Delta}_j) - \Sigma_{jk}| > t\} \cap A_j\right) \\
= \ & \mathbb{P}\left(\{|H^{-1}(\widehat{\tau}_{jk}; \widehat{\Delta}_j) - H^{-1}(F(\Sigma_{jk}; \widehat{\Delta}_j); \widehat{\Delta}_j)| > t\} \cap A_j\right) \\
\le \ & \mathbb{P}\left(\{L_3|\widehat{\tau}_{jk} - H(\Sigma_{jk}; \widehat{\Delta}_j)| > t\} \cap A_j\right) \\
\le \ & \mathbb{P}\left(L_3|\widehat{\tau}_{jk} - H(\Sigma_{jk}; \Delta_j)| > t/2\right) + \mathbb{P}\left(\{L_3|H(\Sigma_{jk}; \widehat{\Delta}_j) - H(\Sigma_{jk}; \Delta_j)| > t/2\} \cap A_j\right) \\
\equiv \ & I_1 + I_2.
\end{aligned} \tag{42}$$

The Hoeffding's inequality for U-statistics yields,

$$I_1 \le 2\exp\left(-\frac{nt^2}{8L_3^2}\right). \tag{43}$$

Recall that $\Phi_{21}(x,y,t) = \partial\Phi_2(x,y,t)/\partial x$, and $\Phi_{22}(x,y,t) = \partial\Phi_2(x,y,t)/\partial y$. As shown in the proof of Theorem 3.1, $\Phi_{21}(x,y,t) \le 1/\sqrt{2\pi}$ and $\phi(x) \le 1/\sqrt{2\pi}$. For $I_2$, we have

$$
\begin{aligned}
& \left|H(\Sigma_{jk};\widehat{\Delta}_j) - H(\Sigma_{jk};\Delta_j)\right| \\
\le\ & 4\left|\Phi_2(\widehat{\Delta}_j,0,\Sigma_{jk}) - \Phi_2(\Delta_j,0,\Sigma_{jk})\right| + 2\left|\Phi(\widehat{\Delta}_j) - \Phi(\Delta_j)\right| \\
\le\ & 4|\Phi_{21}(\xi_1)(\widehat{\Delta}_j - \Delta_j)| + 2|\phi(\xi_2)(\widehat{\Delta}_j - \Delta_j)| \\
\le\ & \frac{6}{\sqrt{2\pi}}\left|\widehat{\Delta}_j - \Delta_j\right|,
\end{aligned}
$$

where $\xi_1$ and $\xi_2$ are the intermediate values from the mean value theorem. Thus, the Hoeffding's inequality yields

$$
I_2 \le \mathbb{P}\left(L_1\left|\frac{1}{n}\sum_{i=1}^n X_{ij} - (1-\Phi(\Delta_j))\right| > \frac{t\sqrt{2\pi}}{12L_3}\right) \le 2\exp\left(-\frac{nt^2\pi}{12^2L_1^2L_3^2}\right). \tag{44}
$$

Combining results (41), (42), (43), (44) and (40), we now obtain

$$
\mathbb{P}\left(\left|H^{-1}(\widehat{\tau}_{jk};\widehat{\Delta}_j) - \Sigma_{jk}\right| > t\right) \le 2\exp\left(-\frac{nt^2}{8L_3^2}\right) + 2\exp\left(-\frac{nt^2\pi}{12^2L_1^2L_3^2}\right) + 2\exp\left(-\frac{M^2n}{2L_1^2}\right).
$$

This implies that

$$
\begin{aligned}
\mathbb{P}\left(\sup_{d_1+1\le j\le d, 1\le k\le d_1}|\widehat{R}_{jk} - \Sigma_{jk}| > t\right) \le\ & 2d_1d_2\exp\left(-\frac{nt^2}{8L_3^2}\right) + 2d_1d_2\exp\left(-\frac{nt^2\pi}{12^2L_1^2L_3^2}\right) \\
& + 2d_1d_2\exp\left(-\frac{M^2n}{2L_1^2}\right),
\end{aligned}
$$

Hence, taking $t = C\sqrt{\log d/n}$ for some constant $C$, $\sup_{d_1+1\le j, 1\le k\le d_1}|\widehat{R}_{jk} - \Sigma_{jk}| \le C\sqrt{\log d/n}$ with probability greater than $1 - d^{-1}$.

□